\documentclass[onefignum,onetabnum]{siamonline190516}

\usepackage{lipsum}
\usepackage{amsfonts}
\usepackage{graphicx}
\usepackage{epstopdf}
\usepackage{algorithmic}
\ifpdf
  \DeclareGraphicsExtensions{.eps,.pdf,.png,.jpg}
\else
  \DeclareGraphicsExtensions{.eps}
\fi
\newcommand{\mat}[1]{\mathbf{#1}}

\usepackage{enumitem}
\setlist[enumerate]{leftmargin=.5in}
\setlist[itemize]{leftmargin=.5in}

\newsiamremark{remark}{Remark}
\newsiamremark{hypothesis}{Hypothesis}
\crefname{hypothesis}{Hypothesis}{Hypotheses}
\newsiamthm{claim}{Claim}

\headers{A Projected Kernelized Stein Discrepancy}{C.Hawkins, Z.Zhang}

\title{A Projected Kernelized Stein Discrepancy\thanks{Submitted to the editors DATE.
\funding{Fill in funding}}}

\author{Cole Hawkins\thanks{Department of Mathematics, University of California Santa Barbara
  (\email{colehawkins@math.ucsb.edu}).}
\and Zheng Zhang\thanks{Department of Electrical and Computer Engineering, University of California Santa Barbara 
  (\email{zhengzhang@ece.ucsb.edu}).}}

\usepackage{amsopn}

\usepackage{amsmath,amsfonts}
\usepackage{algorithmic}
\usepackage{bm}
\usepackage[flushleft]{threeparttable}
\usepackage{tikz}
\usepackage{dirtytalk}
\usetikzlibrary{bayesnet}
\usepackage[nomessages]{fp}
\DeclareMathAlphabet\mathbfcal{OMS}{cmsy}{b}{n}
\newcommand{\ten}[1]{\mathbfcal{#1}} 
\newcommand{\fm}[1]{\mat{U}^{(#1)}}
\newcommand{\ft}[1]{\ten{G}^{(#1)}} 
\newcommand{\rp}[1]{\boldsymbol{\lambda}^{(#1)}} 

\DeclareMathOperator*{\argmin}{arg\,min}
\usepackage{xcolor}

\usepackage{algorithm,algorithmic}
\usepackage[nomessages]{fp}

\newcommand{\thickhline}{\noalign{\hrule height 1.0pt}}

\usepackage{subcaption}
\usepackage{float}

\usepackage[nocompress]{cite}

\begin{document}

\title{Towards Compact Neural Networks via End-to-End Training:\\ A Bayesian Tensor Approach with Automatic Rank Determination \thanks{This work is supported by a Facebook Research Award. CH and ZZ are also supported by NSF CCF-1817037 and DOE DE-SC0021323.} }

%
%
%

\author{Cole Hawkins\thanks{Department of Mathematics, University of California at Santa Barbara
  (\email{colehawkins@math.ucsb.edu}).}
  \and Xing Liu\thanks{Facebook AI Systems Hardware/Software Co-design 
  (\email{xingl@fb.com}).}
\and Zheng Zhang\thanks{Department of Electrical and Computer Engineering, University of California at Santa Barbara 
  (\email{zhengzhang@ece.ucsb.edu}).}}

\markboth{}%

%



\maketitle

\begin{abstract}
Post-training model compression can reduce the inference costs of deep neural networks, but uncompressed training still consumes enormous hardware resources and energy. To enable low-energy training on edge devices, it is highly desirable to directly train a compact neural network from scratch with a low memory cost. Low-rank tensor decomposition is an effective approach to reduce the memory and computing costs of large neural networks. However, directly training low-rank tensorized neural networks is a very challenging task because it is hard to determine a proper tensor rank {\it a priori}, and the tensor rank controls both model complexity and accuracy. This paper presents a novel end-to-end framework for low-rank tensorized training. We first develop a Bayesian model that supports various low-rank tensor formats (e.g., CP, Tucker, tensor train and tensor-train matrix) and reduces neural network parameters with automatic rank determination during training. Then we develop a customized Bayesian solver to train large-scale tensorized neural networks. Our training methods shows orders-of-magnitude parameter reduction and little accuracy loss (or even better accuracy) in the experiments. On a very large deep learning recommendation system with over $4.2\times 10^9$ model parameters, our method can reduce the parameter number to $1.6\times 10^5$ automatically in the training process (i.e., by $2.6\times 10^4$ times) while achieving almost the same accuracy. Code is available at \url{https://github.com/colehawkins/bayesian-tensor-rank-determination}.
\end{abstract}

\section{Introduction}

Despite their success in many applications, deep neural networks are often over-parameterized, requiring extensive computing resources in their training and inference. For instance, the VGG-19 network requires 500M memory~\cite{VGG-16} for image recognition and realistic Deep Learning Recommendation Model (DLRM)~\cite{naumov2019deep} has billions of parameters. It has been a common practice to reduce the size of neural networks before deploying them in various scenarios ranging from cloud services to embedded systems to mobile applications. To reduce hardware cost, numerous techniques have been developed to build {\it compact} models~\cite{alvarez2017compression,hanson1989comparing,lecun1990optimal} after training. Representative approaches include pruning~\cite{lecun1990optimal,neklyudov2017structured}, quantization~\cite{han2015deep,zhou2017incremental,yang2021dynamic}, knowledge distillation~\cite{hinton2015distilling}, and low-rank factorization~\cite{sainath2013low,xue2013restructuring, kim2015compression,lebedev2014speeding,ma2019unified}. Among these techniques, low-rank tensor compression~\cite{kim2015compression,lebedev2014speeding,garipov2016ultimate,he2017wider,cui2019tensor,ma2019tensorized} has achieved possibly the most significant compression, leading to promising reduction of FLOPS and hardware cost~\cite{kim2015compression,deng2019tie}. The recent progress of algorithm/hardware co-design~\cite{zhang2019iccad,deng2019tie} of tensor operations can further reduce the run-time and boost the energy efficiency of tensorized models on edge devices (e.g., on FPGA and ASIC). While post-training compression techniques can reduce the cost of deploying a deep neural network, they cannot reduce the training cost.  

Training consumes far more money, run-time, energy, and hardware resources than inference~\cite{strubell2019energy}. 
Meanwhile, the increasing concerns about data privacy have become a driving force for training on resource-constrained edge devices~\cite{teerapittayanon2017distributed}. The high costs and hardware constraints associated with neural network training motivate us to ask the following question: {\it \say{Is it possible to train a compact neural network model from scratch?}} Both computing and hardware costs may be significantly reduced on various platforms if we can avoid the full-size uncompressed training. While pruning techniques can also be used in training~\cite{neklyudov2017structured,wen2016learning}, they do not necessarily reduce the number of training variables. Low-precision arithmetic~\cite{hubara2017quantized,koster2017flexpoint,gupta2015deep,sun2020ultra} can reduce the cost per parameter during training and inference, but the memory cost reduction is limited to a single order of magnitude even in ultra low-precision 4-bit training~\cite{sun2020ultra}.

\subsection{Contributions}  
\begin{figure}
\centering
    \includegraphics[width=6in]{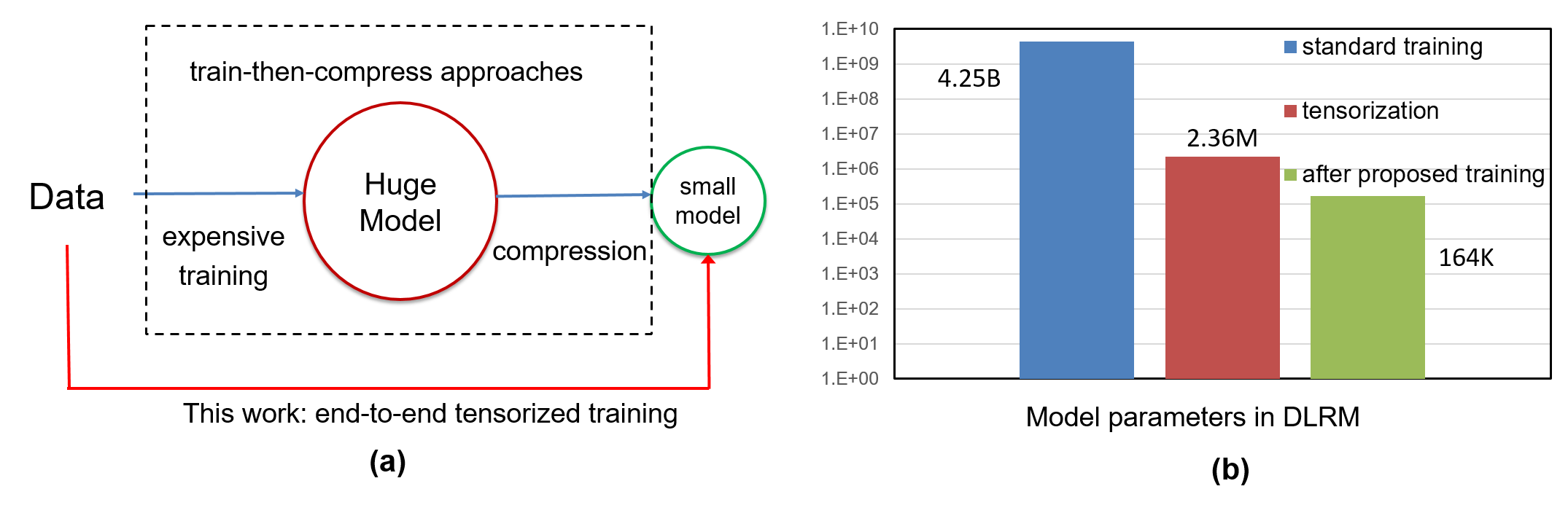}
\caption{(a) Key idea of this work. Conventional train-then-compress approaches have high training costs. In constrast, the proposed end-to-end tensorized training can reduce the training variables significantly and directly produce ultra-compact neural networks. (b) Effectiveness of this approach on a realistic DLRM benchmark. Standard methods train $4.25$ billion variables. Our proposed method only trains $2.36$ million variables, which are further reduced to $164$K in the training process due to the automatic tensor rank determination. }
    \label{fig: overview}
    \vspace{-20pt}
\end{figure}

This paper will present a rank-adaptive end-to-end tensorized training method to generate ultra-compact neural networks directly from scratch. As shown in Fig.~\ref{fig: overview} (a), our method avoids the expensive full-size training in contrast with existing post-training tensor compression methods~\cite{kim2015compression,lebedev2014speeding,garipov2016ultimate,cui2019tensor}. Our method can reduce the training and inference variables by several orders of magnitude, and may achieve further reductions if combined with low-precision numerical operations~\cite{hubara2017quantized,koster2017flexpoint,gupta2015deep,sun2020ultra}. This work can make a great practical impact: it may enable energy-efficient training of medium- or large-size neural networks on edge devices (e.g, embedded GPUs and FPGA), which is impossible to achieve at this moment with existing training methods. Some recent works have studied low-rank tensorized training ~\cite{novikov2015tensorizing,calvi2019compression,khrulkov2019tensorized}, but they fix the tensor ranks before training. It is hard to decide a proper tensor rank parameter {\it a-priori} in practice, therefore one often has to perform extensive combinatorial searches and many training runs until a good rank parameter is found.  

We make the following contributions to achieve efficient one-shot tensorized training:
\begin{itemize}
    \item {\bf A general-purpose rank-adaptive Bayesian tensorized model.} The training cost and model performance are controlled by tensor ranks, which are unknown {\it a priori}.  In order to avoid expensive manual search for tensor ranks required by recent works~\cite{novikov2015tensorizing,calvi2019compression,khrulkov2019tensorized}, we develop a novel Bayesian model to determine both tensor ranks and factors automatically. Existing tensor-based modeling methods are problem-specific and focus on a single tensor format~\cite{zhao2015bayesian,zhao2016bayesian,zhang2018variational,hawkins2019bayesian}. In contrast our work includes all four low-rank tensor formats in common use (CP, Tucker, tensor-train, and tensor-train matrix) and make general advances in low-rank tensor-based modeling. This paper focuses on neural networks, but our method can easily be applied to other tensor problems (e.g., tensor completion, tensor regression and multi-task tensor learning).
    \item {\bf A scalable Stochastic Variational Inference Bayesian solver for the proposed tensorized neural networks.} Training Bayesian tensorized neural networks is expensive, and existing approaches incur high memory and compute requirements. This is because particle-based Bayesian methods require multiple model copies and multiple forward propagations for every training and inference step \cite{hawkins2019bayesian}. Existing mean-field Bayesian tensor completion solvers~\cite{zhao2015bayesian,zhao2016bayesian,zhang2018variational} do not work for tensorized neural networks because of the highly nonlinear forward propagation model in our case. 
    In this work we improve the approximate Bayesian inference method~\cite{hoffman2013stochastic}. Specifically, we observe that directly employing the solver in~\cite{hoffman2013stochastic} causes large gradient variance in our tensorized model. Therefore, we simplify the posterior density of some rank-controlling hyper parameters, and develop an analytical/numerical hybrid approach for the solution update. This customized Bayesian solver infers the unknown tensor factors and tensor ranks of realistic neural networks in a single training run, enabling training and quantifying the uncertainty of extremely large-scale deep learning models that are beyond the capability of existing Bayesian solvers.
    \item {\bf Extensive numerical validations.} We test our algorithms on four benchmarks with model parameters ranging from $4\times 10^5$ to $4.2\times 10^9$. Our method can reduce the training variables by several orders of magnitude with little or even no loss of accuracy. For instance, our method achieves $26,000\times$ parameter reduction when training a large-scale DLRM model as shown in Fig.~\ref{fig: overview} (b). We also compare our methods with existing tensorized neural network methods~\cite{novikov2015tensorizing,garipov2016ultimate,calvi2019compression,khrulkov2019tensorized} including post-training compression and fixed-rank tensorized training, which clearly demonstrates the advantage of our rank-adaptive training method in terms of variable reduction and model accuracy. 
\end{itemize} 
To the best of our knowledge, this work is the first end-to-end Bayesian method that automatically determines the tensor rank in large-scale neural network training (with billions of model parameters) and supports multiple low-rank tensor formats simultaneously. This work will enable energy-efficient and low-cost training of realistic neural networks in resource-constrained scenarios such as internet of things (IoT), robotic systems and mobile phones, as demonstrated by our recent preliminary FPGA prototype for on-device training~\cite{zhang2021fpga}. The Bayesian solution will enable uncertainty quantification of the prediction results, which is important in safety-critical applications such as autonomous driving and medical imaging.


\subsection{Related Work}

There is a massive body of work studying the pruning~\cite{lecun1990optimal,neklyudov2017structured}, quantization~\cite{han2015deep,zhou2017incremental}, knowledge distillation~\cite{hinton2015distilling}, and low-rank compression~\cite{sainath2013low,xue2013restructuring, kim2015compression,lebedev2014speeding} of deep neural networks. This work is most related to the following previous results.

\paragraph{Rank Determination for Linear Tensor Problems} 
Many heuristic methods have been developed to estimate the tensor ranks in tensor factorization and completion. Optimization-based approaches employ a heuristic tensor nuclear norm as the surrogate of tensor rank~\cite{gandy2011tensor, goldfarb2014robust}, but they require expensive regularization on the unfolded tensor. A nice alternative solution is to use Bayesian inference to automatically estimate tensor ranks from observed data~\cite{zhao2015bayesian,zhou2013tensor,guhaniyogi2017bayesian}. Current Bayesian tensor methods solve tensor factorization, completion and regression problems on small-scale data where the observed data is a linear function of the hidden tensor. These problems allow closed-form parameter updates in mean-field Bayesian inference~\cite{zhao2015bayesian,zhou2013tensor,guhaniyogi2017bayesian}. Sampling-based Bayesian methods (i.e. MCMC) require storing thousands of copies of the model, which is not feasible for large neural networks. Because the mean-field variational approach for linear tensor problems~\cite{zhao2015bayesian,zhao2016bayesian} does not work for tensorized neural networks, this paper develops a scalable solver based on stochastic variaitonal inference~\cite{hoffman2013stochastic}. 

\paragraph{Tensorized Neural Networks} Most work uses tensor decomposition to compress pre-trained neural networks. Examples include employing CP and Tucker factorizations to compress convolutional layers~ \cite{lebedev2014speeding,kim2015compression}. In these examples the convolutional filters are already in a tensor form. It is a common practice to reshape the weights in a fully connected layer into a high-order tensor which enables tensor factorization can achieve much higher a compression ratio than matrix factorization on convolution layers~\cite{lebedev2014speeding}. As shown in~\cite{kim2015compression,zhen2019fast}, a neural network compressed by low-rank tensor decomposition can consumes less memory, latency and energy on resource-constrained platforms such as mobile phones. Some recent approaches train low-rank tensorized neural networks~\cite{novikov2015tensorizing,calvi2019compression,hrinchuk2020tensorized} by assuming a low-rank tensorization with a fixed maximum rank. While it is possible to tune the tensor ranks in post-training tensor compression~\cite{lebedev2014speeding,kim2015compression} based on approximation errors, one has to use manual tuning or combinatorial search to determine tensor ranks in existing tensorized training methods~\cite{novikov2015tensorizing,calvi2019compression,hrinchuk2020tensorized}. This has been a major challenge that prevents one-shot training of realistic neural networks on edge devices. Also related to our work are \cite{kossaifi2020factorized} and \cite{kolbeinsson2021tensor}. The work in \cite{kossaifi2020factorized} uses $\ell_1$ regularization to determine CP tensor ranks in a computer vision application but requires multiple hyperparameter tuning runs, which are undesireable in the compressed training setting. The work in \cite{kolbeinsson2021tensor} uses dropout to randomly drop entire tensor ranks as a form of regularization during training. The dropout rate in \cite{kolbeinsson2021tensor}, or rank reduction ratio, is the key hyperparameter that our work determines automatically.

\section{Preliminaries}

\subsection{Tensors and Tensor Decomposition}
\label{subsec:tensor}

\begin{figure}[t]
\centering
  \begin{subfigure}[t]{0.4\textwidth}
  \centering
    \includegraphics[width=.4\textwidth]{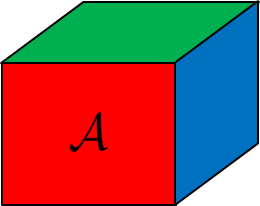}
    \caption{\label{fig: full visual}}
  \end{subfigure}
  \hspace{10pt}
\begin{subfigure}[t]{0.4\textwidth}
  \centering
    \includegraphics[width=1.0\textwidth]{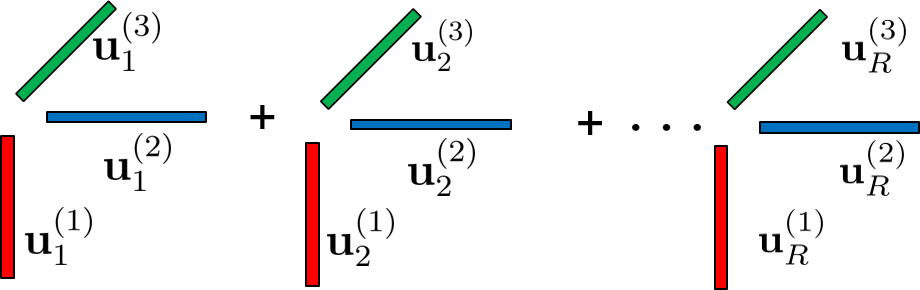}
    \caption{\label{fig: cp visual description}}
  \end{subfigure}
  \\

     \begin{subfigure}[t]{0.4\textwidth}
  \centering
    \includegraphics[width=.9\textwidth]{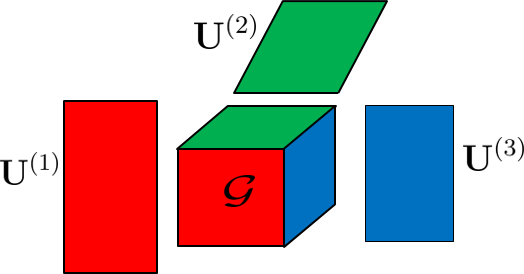}
    \caption{\label{fig: tucker visual description}}
  \end{subfigure}
    \hspace{10pt}
     \begin{subfigure}[t]{0.4\textwidth}
  \centering
    \includegraphics[width=1.0\textwidth]{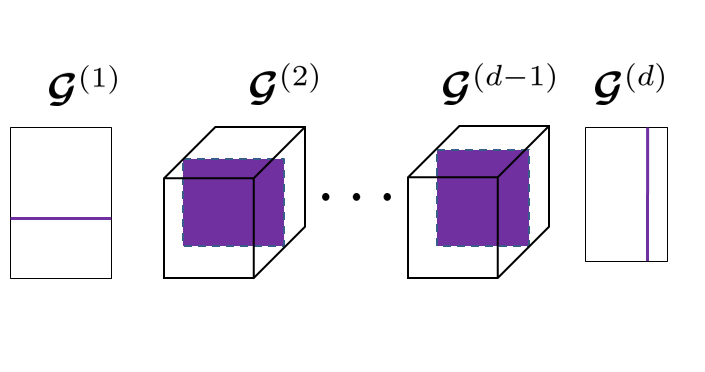}
    \caption{\label{fig: tt visual description}}
  \end{subfigure}
  \caption{(a): An order-3 tensor, (b) and (c): representations in CP and Tucker formats respectively, where low-rank factors are color-coded to indicate the corresponding modes. (d): TT representation of an order-$d$ tensor, where the purple lines and squares indicate $\ten{G}^{(n)}(:,i_n,:)$, which is the $i_n$-th slice of the TT core $\ten{G}^{(n)}$ obtained by fixing its second index. \label{fig: visual formats}}
\end{figure}

This paper uses lower-case letters (e.g., $a$) to denote scalars, bold lowercase letters (e.g., $\mat{a}$) to represent vectors, bold uppercase letters (e.g., $\mat{A}$) to represent matrices, and bold calligraphic letters (e.g., $\ten{A}$) to denote tensors. A tensor is a generalization of a matrix, or a multi-way data array. An order-$d$ tensor is a $d$-way data array $\ten{A}\in \mathbb{R}^{I_1 \times I_2 \times \dots \times I_d}$, where $I_n$ is the size of mode $n$. The $(i_1, i_2, \cdots, i_d)$-th element of $\ten{A}$ is denoted as $a_{i_1i_2 \cdots i_d}$. An order-$3$ tensor is shown in Fig.~\ref{fig: visual formats} (a).                                                                                         

\begin{definition}
The  mode-$n$ product of a tensor $\ten{A}\in\mathbb{R}^{I_1\times\dots \times I_{n} \times \dots\times I_d}$ with a matrix $\mat{U}\in\mathbb{R}^{J\times I_n}$ is 
\begin{equation}
    \label{eq: mode n product}
    \begin{split}
    \ten{B} &= \ten{A}\times_n \mat{U} \Longleftrightarrow
    b_{i_1\dots i_{n-1} j i_{n+1} \dots i_d}=\sum_{i_n=1}^{I_n} a_{i_1  \dots i_d}  u_{j i_n}  .
    \end{split}
\end{equation}
\end{definition}
The result is still a $d$-dimensional tensor $\ten{B}$, but the mode-$n$ size becomes $J$. In the special case $J=1$, the $n$-th mode diminishes and $\ten{B}$ becomes an order-$d-1$ tensor.

A tensor has a massive number of entries if $d$ is large. This causes a high cost in both computing and storage. Fortunately, many practical tensors have a low-rank structure, and this property can be exploited to reduce the cost dramatically. 

\begin{definition}
A $d$-way tensor $\ten{A} \in \mathbb{R}^{I_1\times \cdots \times I_d}$ is rank-1 if it can be written as a single outer product of $d$ vectors
\begin{equation}
\ten{A} =\mat{u}^{(1)} \circ \dots \circ \mat{u}^{(d)}, \; {\text{with}} \; \mat{u}^{(n)} \in\mathbb{R}^{I_n} \; \text{for}\; n=1,\cdots, d. \nonumber
\end{equation}
\end{definition}
Each element of $\ten{A}$ is $a_{i_1 i_2 \cdots i_d}=\prod \limits_{n=1}^d u_{i_n}^{(n)}$,
where $u_{i_n}^{(n)}$ is the $i_n$-th element of the vector $\mat{u}^{(n)}$. 

A rank-1 tensor can be stored with only $d$ vectors. Most tensors are not rank-1, but many can be well-approximated via tensor decomposition~\cite{kolda2009tensor} if their ranks are low. We will use the following four tensor decomposition formats to reduce the parameters of neural networks.

\begin{definition}\label{Def: CP}
The CP factorization~\cite{carroll1970analysis,harshman1994parafac} expresses tensor $\ten{A}$ as the sum of multiple rank-1 tensors:
\begin{equation}
\ten{A} = \sum_{j=1}^R \mat{u}_j^{(1)} \circ \mat{u}_j^{(2)} \dots \circ \mat{u}_j^{(d)}.
\end{equation}
Here $\circ$ denotes an outer product operator. The minimal integer $R$ that ensures the equality is called the {\bf CP rank} of $\ten{A}$. To simplify notation we collect the rank-1 terms of the $n$-th mode into a factor matrix $ \fm{n} \in \mathbb{R}^{I_n \times R}$ with $\fm{n}(:,j) = \mat{u}_{j}^{(n)}$. A rank-$R$ CP factorization can be described with $d$ factor matrices $\{ \mat{U}^{(n)}\}_{n=1}^d$ using $R\sum_n I_n$ parameters.
\end{definition}

\begin{definition}\label{Def: Tucker}
The Tucker factorization~\cite{tucker1966some} expresses a $d$-way tensor $\ten{A}$ as a series of mode-$n$ products:
\begin{equation}
\label{eq:Tucker}
\ten{A} = \ten{G}\times_1 \fm{1}\times_2 \dots \times_d \fm{d}.
\end{equation}
Here $\ten{G}\in\mathbb{R}^{R_1\times \dots\times R_d}$ is a small core tensor, and $\fm{n}\in\mathbb{R}^{I_n\times {R}_n}$ is a factor matrix for the $n$-th mode. The {\bf Tucker rank} is the tuple $(R_1,\dots,R_d)$. A Tucker factorization with ranks $R_n=R$ requires $R^d+R\sum_n I_n$ parameters.
 \end{definition}

\begin{definition}
\label{def: tensor train}
The tensor-train (TT) factorization~\cite{oseledets2011tensor} expresses a $d$-way tensor $\ten{A}$ as a collection of matrix products:
\begin{equation}
\label{eq:TT}
a_{i_1 i_2 \dots i_d} = \ft{1}(:,i_1,:)\ft{2}(:,i_2,:)\dots \ft{d}(:,i_d,:).
\end{equation}
Each  TT-core $\ft{n}\in \mathbb{R}^{R_{n-1}\times I_n \times R_{n}}$ is an order-$3$ tensor. The tuple $(R_0,R_1,\dots,R_d)$ is the {\bf TT-rank} and $R_0=R_d=1$.
\end{definition}
The TT format uses $\sum_n R_{n-1}I_nR_{n}$ parameters in total and leads to more expressive interactions than the CP format.

Let $\mat{A}\in\mathbb{R}^{I\times J}$ be a matrix. We assume that $I$ and $J$ can be factored as follows:
\begin{equation}
\label{eq: TTM dimension factorization}
    I=\prod_{n=1}^d I_n,J=\prod_{n=1}^d J_n.
\end{equation}
We can reshape $\mat{A}$ into a tensor $\ten{A}$ with dimensions $I_1\times\dots\times I_d\times J_1\times\dots\times J_d$, such that the $(i,j)$-th element of $\mat{A}$ uniquely corresonds to the $(i_1, i_2, \cdots, i_d, j_1, j_2, \cdots, j_d)$-th element of $\ten{A}$. The TT decomposition can extended to compress the resulting order-$2d$ tensor as follows.

\begin{definition}
\label{def: tensor train matrix}
The tensor-train matrix (TTM) factorization expresses an order-$2d$ tensor $\ten{A}$ as $d$ matrix products: 
\begin{equation}
\label{eq: ttm factorization}
a_{i_1\dots i_d j_1 \dots j_d} = \ft{1}(:,i_1,j_1,:)\ft{2}(:,i_2,j_2,:)\dots \ft{d}(:,i_d,j_d,:).
\end{equation}
Each TT-core $\ft{n}\in \mathbb{R}^{R_{n-1}\times I_n \times J_n \times R_{n}}$ is an order $4$ tensor. The tuple $(R_0,R_1,R_2,\dots,R_d)$ is the {\bf TT-rank} and as before $R_0=R_d=1$. This TTM factorization requires $\sum_n R_{n-1}I_n J_nR_{n}$ parameters to represent $\ten{A}$. \end{definition}

We provide a visual representation of the CP, Tucker, and TT formats in Fig.~\ref{fig: visual formats} (b) -- (d).

\subsection{Tensorized Neural Networks} 

A deep neural network can be written as 
\begin{equation}
\label{eq: neural network}
    \mat{y}=\mat{h}(\mat{x})=\mat{g}_L\left( \mat{g}_{L-1} \left( \cdots \mat{g}_1(\mat{x}) \right)\right)
\end{equation}
where $\mat{x}$ is an input data sample and $\mat{y}$ is an output label. Here $\mat{g}_k(\mat{z})=\sigma(\mat{W}_k \mat{z} +\mat{b}_k)$ represents layer $k$, where $\sigma$ is a nonlinear activation function, $\mat{W}_k$ and $ \mat{b}_k$ are the weights and bias, respectively.  Considering parameter dependence, we can re-write \eqref{eq: neural network} as
\begin{equation}
\label{eq: tensor neural network}
    \mat{y}=\mat{h}(\mat{x}\;| \;\{\mat{W}_k, \mat{b}_k \}_{k=1}^L).
\end{equation}
In a convolutional layer $\mat{W}_k$ should be replaced with tensor $\ten{W}_k$. In modern neural networks, $\{\mat{W}_k\}_{k=1}^L$ contain millions to billions of parameters, which cause huge challenges in training and inference on various hardware platforms. A promising solution is to generate a compact neural network via low-rank tensor compression~\cite{lebedev2014speeding,novikov2015tensorizing,garipov2016ultimate} as follows:
\begin{itemize}
    \item {\bf Folding to high-order tensors.} A weight matrix $\mat{W}\in\mathbb{R}^{I\times J}$ can be folded into an order-$d$ tensor $\ten{A}\in\mathbb{R}^{I_1\times\dots\times I_d}$ where $IJ=\prod_n I_n$. We can also fold $\mat{W}$ to an order-$2d$ tensor $\ten{A} \in \mathbb{R}^ {I_1 \times \cdots \times I_d \times J_1 \times \cdots \times J_d }$ such that $w_{ij}=a_{i_1\cdots i_d j_1 \cdots j_d}$. While a convolution filter is already a tensor, we can reshape it to a higher-order tensor with reduced mode sizes.  
    \item {\bf Low-rank tensor compression.} After folding $\mat{W}$ into a higher-order tensor $\ten{A}$, one can employ low-rank tensor compression to reduce the number of parameters. Either the CP, Tucker, TT or TTM factorization can be applied~\cite{lebedev2014speeding,kim2015compression,garipov2016ultimate}. 
\end{itemize}

Assume that $\boldsymbol{\Phi}_k$ includes all low-rank tensor factors required to represent $\mat{W}_k$ . Considering the dependence of $\mat{W}_k$ on $\boldsymbol{\Phi}_k$, we can now write \eqref{eq: tensor neural network} as
\begin{align}
\label{eq: low rank tensor NN}
    \mat{y}=&\mat{h}\left(\mat{x}\;| \;\{\mat{W}_k \left( \boldsymbol{\Phi}_k \right), \mat{b}_k \}_{k=1}^L\right)= \mat{f} \left(\mat{x}\;| \; \boldsymbol{\Psi}\right), \; {\text{with}}\; \boldsymbol{\Psi}=\{ \boldsymbol{\Phi}_k, \mat{b}_k\}_{k=1}^L.
\end{align}
Here $\boldsymbol{\Psi}$ include all tensor factors and bias vectors in a tensorized neural network. The number of variables in $\boldsymbol{\Psi}$ is often orders-of-magnitude smaller than that in the original model~\eqref{eq: tensor neural network}.

Please note the following:
\begin{itemize}
    \item The tensor factors in $\boldsymbol{\Phi}_k$ depend on the tensor format we choose. In CP format, $\boldsymbol{\Phi}_k$ includes $d$ matrix factors; in Tucker format, $\boldsymbol{\Phi}_k$ includes $d$ factor matrices and a small order-$d$ core tensor as shown in \eqref{eq:Tucker}; when the TT or TTM format is used, $\boldsymbol{\Phi}_k$ includes $d$ order-$3$ or order-$4$ TT cores shown in \eqref{eq:TT} and \eqref{eq: ttm factorization} respectively. 
    \item The number of variables in each $\boldsymbol{\Phi}_k$ depends on the tensor ranks used in the compression. A higher tensor rank leads to higher expressive power but a lower compression ratio. In existing approaches, it is hard to select a proper tensor rank {\it a-priori}.  
    
\end{itemize}

Two main approaches exist to produce low-rank tensorized neural networks. The first approach trains an uncompressed neural network $\mat{h}$ and then performs tensor factorization on each of the weights $\{\mat{W}_k\}_{k=1}^L$. This train-then-compress approach suffers from two drawbacks: 
\begin{itemize}
    \item {\bf High training costs.} The uncompressed training consumes a huge amount of memory, run-time, and energy on a hardware platform. 
    \item {\bf Lower accuracy.} The subsequent tensor compression causes accuracy loss, which becomes significant when the compression ratio is high.
    \end{itemize}
The second approach is fixed-rank tensorized training. In this approach the user pre-specifies the tensor rank and trains low-rank tensor factors of weight parameters. This approach avoids the compute and memory requirements of uncompressed training but requires that the user manually select a good rank {\it a-priori}. This approach usually requires multiple training runs to select the rank. In addition a user-specified rank may achieve suboptimal compression.

\section{Bayesian Low-Rank Tensorized Model}
In this work, we plan to develop a tensorized training method that can automatically determine the tensor ranks in the training process. This method requires only one training run and avoids the high cost of uncompressed training. Bayesian methods have been employed for tensor completion and factorization~\cite{zhou2013tensor,zhao2015bayesian,zhao2016bayesian}, where the observed data is a linear function of tensor elements. However, existing Bayesian tensor solvers do not work for tensorized neural networks due to the nonlinear forward model and large number of unknown variables.

\subsection{High-Level Bayesian Formulation}

We first describe a general-purpose Bayesian model for training low-rank tensorized neural networks. For notational convenience we assume that our neural network $\mat{f}$ has one nonlinear layer, and that its weight matrix $\mat{W}$ is folded to a single tensor $\ten{A}$. Extending our method to general multi-layer cases with multiple tensors is straightforward, and we will report results on general multi-layer models in Section~\ref{sec:experiments}.

Given a training data set ${\cal D}$, our goal is to determine the unknown low-rank factors $\boldsymbol\Phi$ for $\ten{A}$, the associated tensor ranks, and the bias vector $\mat{b}$. We introduce hyper parameters $\boldsymbol \Lambda$ to control the tensor ranks and model complexity. Our posterior distribution is
\begin{equation}
    \label{eq: posterior}
    p(\boldsymbol\Psi,\boldsymbol\Lambda|\mathcal{D})=\frac{p(\mathcal{D}|\boldsymbol\Psi)p(\boldsymbol\Psi,\boldsymbol\Lambda)}{p(\mathcal{D})}, \; \text{with}\; \boldsymbol{\Psi}= \{ \boldsymbol{\Phi}, \mat{b} \}.
\end{equation}
Here $p(\mathcal{D}|\boldsymbol\Psi)$ is the model likelihood, $p(\boldsymbol\Psi,\boldsymbol\Lambda)$ is the joint prior and $p(\mathcal{D})$ is the model evidence
\begin{equation}
    \label{eq: model evidence}
p(\mathcal{D}) = \int_{\boldsymbol\Psi,\boldsymbol\Lambda}p(\mathcal{D}|\boldsymbol\Psi)p(\boldsymbol\Psi,\boldsymbol\Lambda)d\boldsymbol\Psi d\boldsymbol\Lambda.
\end{equation}

The likelihood and joint prior are specified below:
\begin{itemize}
    \item {\bf Likelihood function:}  $p(\mathcal{D}|\boldsymbol\Psi$) and data $\mathcal{D}$ are determined by a forward propagation model. Let $(\mat{x},\mat{y})\in \mathcal{D}$ be a training sample where $\mat{x}$ is the neural network input and $\mat{y}$ is the associated true label. The multinomial likelihood function for a neural network classifier with $C$ potential classes is
\begin{equation}
    \label{eq : likelihood}
    \begin{split}
    p(\mathcal{D}|\boldsymbol\Psi) &\propto \prod_{(\mat{x},\mat{y})\in \mathcal{D}} \prod_{c=1}^C \mat{f}(\mat{x}|\boldsymbol\Psi)_c^{y_c}.
    \end{split}
\end{equation}
where $y_c$ is the correct class label. Here $\mat{f}$ is the forward propagation model in \eqref{eq: low rank tensor NN} which is conditioned on the given low-rank tensor factors and bias vectors. We omit the multinomial distribution constant of proportionality for simplicity.

\item {\bf Joint Prior:} We place an independent prior over the low-rank tensor factors and the bias term. 
We choose a weak normal prior for the bias term:
\begin{equation}
    \label{eq: bias prior}
\begin{split}
    p(\boldsymbol\Psi,\boldsymbol\Lambda)&=p(\mat{b})p(\boldsymbol\Phi,\boldsymbol\Lambda),\;\;
    p(\mat{b}) \propto \prod_i \frac{1}{\sigma_0^2}\exp\left(-\frac{b_i^2}{2\sigma_0^2}\right).
\end{split}
\end{equation}
\end{itemize}
Here $p(\boldsymbol\Phi,\boldsymbol\Lambda)$ is the joint prior for tensor factors $\boldsymbol\Phi$ and hyper parameters $\boldsymbol\Lambda$. The design of $p(\boldsymbol\Phi,\boldsymbol\Lambda)$ depends on the tensor format we choose, which will be explained in Section~\ref{subsec:low-rank prior} \& \ref{subsec:hyper priors}.

\subsection{Tensor Factor Priors}
\label{subsec:low-rank prior}

Proper priors should be chosen in order to automatically shrink tensor ranks in the training process. Here we will specify the joint prior $p(\boldsymbol\Phi,\boldsymbol\Lambda)$ for the four tensor formats described in Section~\ref{subsec:tensor}: CP, Tucker, TT and TTM.

Firstly we specify the general form of $p(\boldsymbol\Phi,\boldsymbol\Lambda)$. For the CP format, we initialize each factor $\mat{U}^{(n)}$ as a matrix with $R$ columns. Assume that $R$ is larger than the actual rank $r$, and all factors shrink to $r$ columns in the training process. All CP factors have the same maximum rank (column number) so we use a single vector $\boldsymbol\Lambda=\boldsymbol{\lambda} \in \mathbb{R}^R$ to control the rank. The tensor rank in Tucker, TT or TTM format is a vector, and the rank associated with each mode can be different. Therefore, we require a collection of vectors $\boldsymbol\Lambda=\{\rp{n}\}_{n=1}^d$ to control the ranks of each mode individually. Here $\rp{n} \in \mathbb{R}^{R_n}$, and the ``maximum rank" $R_n$ exceeds the ``actual rank" $r_n$ of mode $n$. As a result, we introduce the general form 
\begin{equation}
    \label{eq: general low rank prior}
    p(\boldsymbol\Phi,\boldsymbol{\Lambda})= \begin{cases} p(\boldsymbol\Phi|\boldsymbol{\lambda})p(\boldsymbol{\lambda}) \text{ for CP format } \\
     p(\boldsymbol\Phi|\{\rp{n}\})\prod \limits_{n=1}^d p(\rp{n}) \text{ for Tucker, TT \& TTM formats} \\
     \end{cases}
\end{equation}
where the prior distribution(s) on $\boldsymbol{\lambda}$ or $\{\rp{n}\}_{n=1}^d$ enforce(s) rank reduction.  

\begin{figure}[t]
\centering
  \begin{subfigure}[b]{0.23\textwidth}
  \centering
  \raisebox{0cm}{\includegraphics[width=\textwidth]{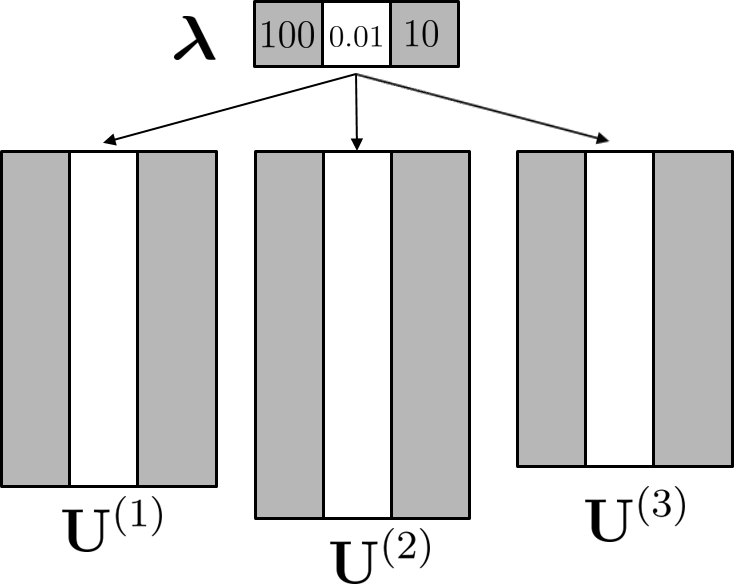}}
  \caption{\label{fig: cp rank shriankge}} 
  \end{subfigure}
  \hspace{35pt}
   \begin{subfigure}[b]{0.23\textwidth}
  \centering
  \raisebox{0cm}{\includegraphics[width=\textwidth]{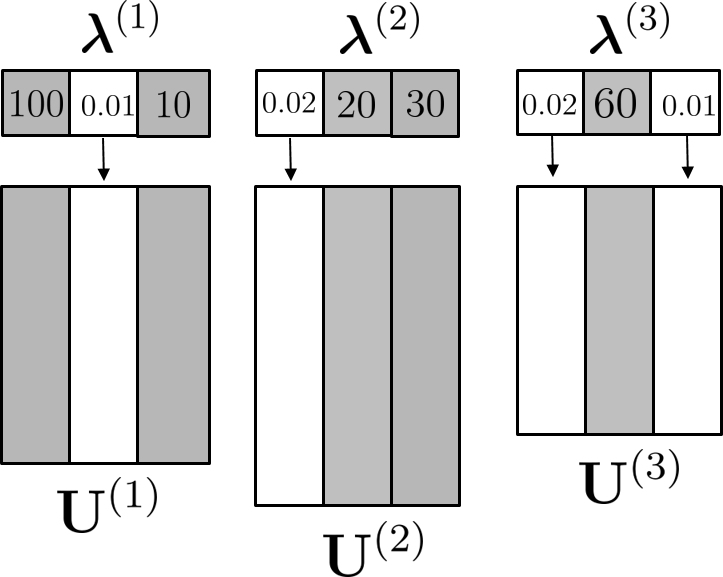}}    
    \caption{\label{fig: tucker rank shrinkage}} 
  \end{subfigure}
 \hspace{35pt}
   \begin{subfigure}[b]{0.18\textwidth}
  \centering
\raisebox{0cm}{\includegraphics[width=\textwidth]{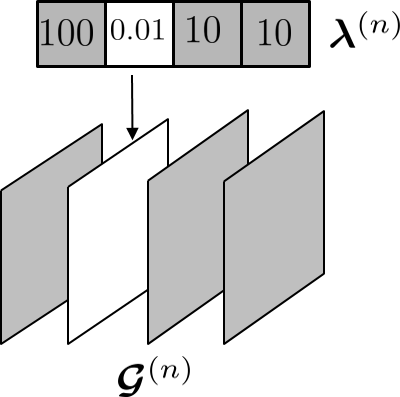}}
  \caption{\label{fig: TT/TTM rank shrinkage}} 
  \end{subfigure}
  \caption{(a) For the CP prior, if one element of $\boldsymbol\lambda$ is small, one column is removed from every factor matrix. (b) For the Tucker prior, if one element of $\boldsymbol\lambda^{(n)}$ is small then one column of $\fm{n}$ shrinks to zero. (c) For the TT prior, if one element of $\boldsymbol\lambda^{(n)}$ is small then one slice of $\ft{n}$ shrinks to zero. The columns/slices to be removed are marked in white.\label{fig: rank shrinkage}}
\end{figure}

Next we specify the tensor factor priors $p(\boldsymbol\Phi|\boldsymbol{\lambda})$ or $p(\boldsymbol\Phi|\{\rp{n}\})$ for each tensor format, and we defer the prior on $\boldsymbol{\lambda}$ and $\{\rp{n}\}_{n=1}^d$ to Section~\ref{subsec:hyper priors}.
\begin{itemize}
\item {\bf CP Format:} The CP tensor factors are $d$ matrices  $\boldsymbol\Phi=\{\mat{U}^{(n)}\}_{n=1}^d$. We assign a Gaussian prior with controllable variance to each element of each factor matrix $\mat{U}^{(n)}$:
\begin{equation}
    \label{eq: cp factor prior}
    \begin{split}
    p(\boldsymbol\Phi,\boldsymbol\Lambda) &= p\left(\boldsymbol{\lambda} \right)\prod_{n} p\left(\fm{n}|\boldsymbol{\lambda} \right), \;\;        
    p\left(\mat{U}^{(n)}\:|\:\boldsymbol{\lambda}\right) = \prod_{i,j} \mathcal{N}\left( u^{(n)}_{i j}\:|\:0, {\lambda}_j\right).
    \end{split}
\end{equation}
Here $u^{(n)}_{i j}$ is the $(i,j)$-th element of $\mat{U}^{(n)}$. Each entry of $\boldsymbol{\lambda}$ controls one column of each factor matrix. If a single entry $\lambda_j$ approaches zero, then the prior mean and prior variance of $u^{(n)}_{ij}$ are both close to zero for all row indices $i \in [1, I_n]$ and mode indices $ n \in [1, d]$. This encourages the whole $j$-th column of $\mat{U}^{(n)}$ to shrink to zero, leading to a rank reduction. The vector $\boldsymbol{\lambda}$ is shared across all modes, therefore it will shrink the same column of all CP factor matrices simultaneously, as shown in Fig.~\ref{fig: rank shrinkage} (a).  

\item {\bf Tucker Format:} A Tucker factorization includes a core tensor and $d$ factor matrices, therefore  $\boldsymbol\Phi=\{\ten{G},\{\fm{n}\}_{n=1}^d\}$. We also assign each factor matrix $\mat{U}^{(n)}$ with a variance-tunable Gaussian distribution. A Tucker model has $d$ separate rank parameters $(r_1,\dots,r_d)$ to determine, one per factor matrix as shown in Fig.~\ref{fig: rank shrinkage} (b). Furthermore, the factor matrices and core tensor are handled separately. Therefore, we propose the following prior distributions:
\begin{equation}
    \label{eq: tucker factor prior}
    \begin{split}
    p(\boldsymbol\Phi,\boldsymbol\Lambda) &= p(\ten{G}) \prod_{n} p\left(\fm{n}|\rp{n} \right)p\left(\rp{n} \right),\;\;        
    p\left(\fm{n}\:|\:\rp{n}\right) = \prod_{i,j} \mathcal{N}\left( u^{(n)}_{ij}\:|\:0, \lambda^{(n)}_j\right). \nonumber
    \end{split} 
\end{equation}
We use $d$ independent rank controlling vectors $\{\rp{n}\}_{n=1}^d$ to control the prior variances of different factor matrices separately. The $j$-th element of $\rp{n}$ controls the $j$-th column of factor matrix $\fm{n}$. Therefore $\rp{n}$ controls $r_n$, the $n$-th entry of the Tucker rank. We place a weak normal prior over the entries of the core tensor $\ten{G}$:
\begin{equation}
    \label{eq: tucker core prior}
    p\left(\ten{G}\right) = \prod_{i_1,\dots,i_d} \mathcal{N}\left( g_{i_1\dots i_d}\:|\:0, \sigma_0\right).
\end{equation}
We make this choice to simplify parameter inference compared to the alternative of placing low-rank priors on both of the core tensor and the factor matrices.

\item {\bf Tensor-Train (TT) Format:} A TT factorization has $d$ order-3 TT cores, therefore $\boldsymbol\Phi=\{\ft{n}\}_{n=1}^d$. The TT format requires a more complicated prior because each TT core $\ft{n}\in \mathbb{R}^{r_{n-1} \times I_n \times r_n}$ depends on two rank parameters $r_{n-1}$ and $r_n$. In order to automatically determine the TT rank, we choose $R_n>r_n$, and initialize the $n$-th TT core with size $R_{n-1} \times I_n \times R_n$. The prior density of all TT cores are given as
\begin{equation}
    \label{eq: tt factor prior}
    \begin{split}
    p(\boldsymbol\Phi,\boldsymbol\Lambda)  &
    = p\left(\ft{d} |\rp{d-1}\right)  \prod_{1\leq n\leq d-1} p\left(\ft{n}|\rp{n} \right)p\left(\rp{n} \right),\\        
    p\left(\ft{n}\:|\:\rp{n}\right) &= \prod_{i,j,k} \mathcal{N}\left( g^{(n)}_{ijk}\:|\:0, \lambda^{(n)}_k\right) \; \text{for} \; n \in [1, d-1], \; \; \\
    p\left(\ft{d}\:|\:\rp{d-1}\right)& = \prod_{i,j,k} \mathcal{N}\left( g^{(d)}_{ijk}\:|\:0, \lambda^{(d-1)}_i\right).
    \end{split}
\end{equation}
We introduce a vector $\rp{n} \in \mathbb{R}^{R_n}$ to control the actual rank $r_{n}$ for mode $1$ to $d-1$. As shown in Fig.~\ref{fig: rank shrinkage} (c), the $k$-th element of $\rp{n}$ (i.e., $\lambda_k^{(n)}$) controls the prior variance of a slice $\ten{G}^{(n)}(:,:,k)$. If $\lambda_k^{(n)}$ is small, the whole slice $\ten{G}^{(n)}(:,:,k)$ is close to zero, leading to a rank reduction in the $n$-th mode. Parameter $\rp{d-1}$  controls two separate cores. This prevents any rank parameters from overlapping and it simplifies posterior inference.

\item {\bf Tensor-Train Matrix (TTM) Format:} Similar to the TT format, a TTM decomposition also has $d$ core tensors, therefore  $\boldsymbol\Phi=\{\ft{n}\}_{n=1}^d$. The only difference is that each $\ten{G}^{(n)}$ is an order-$4$ tensor, which is initalized with a size $R_{n-1} \times I_n \times J_n \times R_{n}$ in our Bayesian model. The prior for the TTM low-rank factors is
\begin{equation}
    \label{eq: ttm factor prior}
    \begin{split}
    p(\boldsymbol\Phi,\boldsymbol\Lambda) & =
    p\left(\ft{d} |\rp{d-1}\right)\prod_{1\leq n \leq d-1} p\left(\ft{n}|\rp{n} \right)p\left(\rp{n} \right),\\        
    p\left(\ft{n}\:|\:\rp{n}\right) &= \prod_{i,j,k,l} \mathcal{N}\left( g^{(n)}_{ijkl}\:|\:0, \lambda^{(n)}_l\right), \; \text{for}\; n\in [1,d-1], \\
    p\left(\ft{d}\:|\:\rp{d-1}\right) &  = \prod_{i,j,k,l} \mathcal{N}\left( g^{(d)}_{ijkl}\:|\:0, \lambda^{(d-1)}_i\right).
    \end{split}
\end{equation}
This prior very similar to that of TT format. We use a vector parameter $\rp{n}$ to control the actual rank $r_n$ of the $n$-th mode for $n\in [1,d-1]$, and $\rp{d-1}$ is shared among $\ten{G}^{(d)}$ and $\ten{G}^{(d-1)}$. 
\end{itemize}

\subsection{Rank-Shrinking Hyper-Parameter Priors}
\label{subsec:hyper priors}
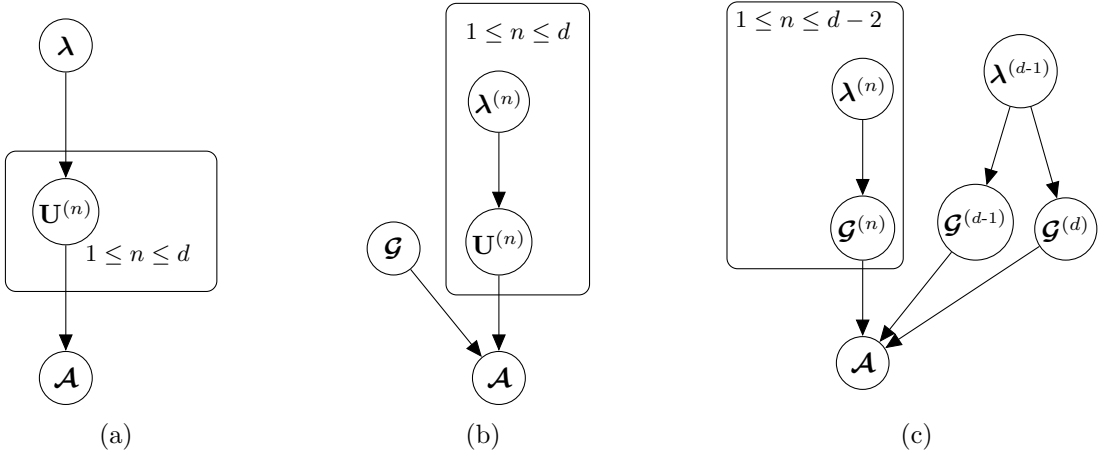
\begin{figure}[t]
  \begin{subfigure}[b]{0.25\textwidth}
  \centering
  \begin{tikzpicture}

  \node[latent]                               (A) {$\ten{A}$};
  \node[latent, above=of A, xshift=0.0cm,yshift=0.4cm] (U) {$\fm{n}$};
  \node[latent, above=of U, xshift=0.0cm,yshift=0.4cm]  (lambda) {$\boldsymbol{\lambda}$};

  \edge {lambda} {U} ; %
  \edge {U} {A} ; %

  \tikzset{plate caption/.append style={below right=25pt and 20pt of #1.north west}}
  \plate  [inner sep=0.35cm] {uplate} {(U)} {\footnotesize{$1\leq n\leq d$}};

\end{tikzpicture}
  \caption{\label{fig: cp graphical}} 
  \end{subfigure}
  \quad\quad
   \begin{subfigure}[b]{0.26\textwidth}
  \centering
    \begin{tikzpicture}

  \node[latent]                               (A) {$\ten{A}$};
  \node[latent, above=of A, xshift=0.0cm] (U) {$\fm{n}$};
  \node[latent, above=of U, xshift=0.0cm]  (lambda) {$\rp{n}$};
  \node[latent, above=of A, xshift=-1.4cm]  (core) {$\ten{G}$};

  \edge {lambda} {U} ; %
  \edge {U} {A} ; %
  \edge {core} {A} ; %

  \tikzset{plate caption/.append style={above right=10pt and 0pt of #1.north west}}
  \plate  [inner sep=0.25cm] {uplate} {(U) (lambda)} {\footnotesize{$1\leq n\leq d$}};

\end{tikzpicture}
  \caption{\label{fig: tucker graphical}} 
  \end{subfigure}
\hspace{0.2in}
   \begin{subfigure}[b]{0.38\textwidth}
  \centering
 \raisebox{0.2cm}{\begin{tikzpicture}
[latent/.append style={minimum size=0.7cm}]
  \node[latent]                               (A) {$\ten{A}$};
  \node[latent, above=of A, xshift=0.0cm] (U) {$\ft{n}$};
  \node[latent, above=of U, xshift=0.0cm]  (lambda) {$\rp{n}$};
  \node[latent, above=of A, xshift=1.5cm]  (penultimate_factor) {$\ft{\text{\tiny{$d$-1}}}$};
  \node[latent, above=of A, xshift=2.7cm]  (last_factor) {$\ft{d}$};
  \node[latent, above=of penultimate_factor, xshift=0.6cm] (last_lambda) {$\rp{\text{\tiny{$d$-1}}}$};
  \edge {lambda} {U} ; %
  \edge {last_lambda} {penultimate_factor,last_factor} ; %
  \edge {U,penultimate_factor,last_factor} {A} ; %
  \tikzset{plate caption/.append style={above left=10pt and -20pt of #1.north west}}
  \plate  [inner sep=0.1cm] {uplate} {(U) (lambda)} {\footnotesize{$1\leq n\leq d-2$}};
\end{tikzpicture}
}
  \caption{\label{fig: TT/TTM graphical}} 
  \end{subfigure}
  \caption{(a) CP graphical model (b) Tucker graphical model (c) TT/TTM graphical model. \label{fig: prior description}}
\end{figure}
 
To complete the setup of the full Bayesian model~\eqref{eq: posterior}, we still need to specify the prior of rank-control hyper parameters $\boldsymbol \Lambda=\boldsymbol{\lambda}$ (for CP) or $\boldsymbol \Lambda=\{\rp{n}\}_{n=1}^d$ (for Tucker, TT and TTM). Since small elements in $\boldsymbol{\lambda}$ and $\rp{n}$ lead to rank reductions in the tensor models, we choose two hyper-prior densities that place high probability near zero. We focus our notation in this subsection on the CP model for simplicity. 

\begin{figure}
    \centering
\begin{subfigure}{0.48\textwidth}
    \includegraphics[width=\textwidth]{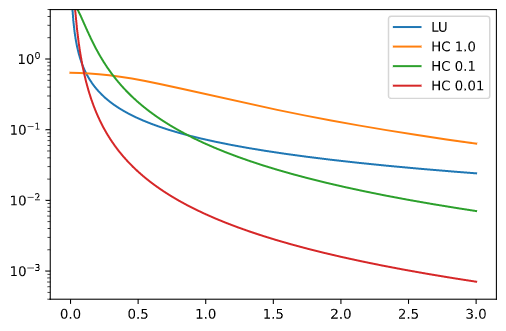}
    \caption{\label{fig: hyperprior}}
\end{subfigure}
\begin{subfigure}{0.48\textwidth}
    \includegraphics[width=\textwidth]{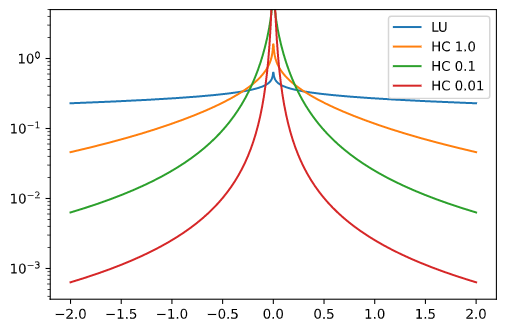}
    \caption{\label{fig: marginal prior}}
\end{subfigure}
    \caption{(a) Comparison of the probability density functions of the Log-Uniform and Half-Cauchy hyperprior on $\lambda_j$. Several values of the Half-Cauchy scale parameter $\eta$ are given. (b) Comparison of the probability density functions of the corresponding marginal prior on the low-rank tensor factor entry $u_{ij}^{(n)}$.}
    \label{fig: pdf}
\end{figure}

We consider two choices of prior on the hyper parameter $\boldsymbol{\lambda}$: the Half-Cauchy with scale parameter $\eta$ and the improper Log-Uniform on $(0,\infty)$: 
\begin{equation}
\label{eq: prior choices}
    \begin{split}
        p(\boldsymbol{\lambda}) &= \prod_{i=1}^R p(\lambda_i),\quad
        \text{with}\; p(\lambda_i) =\begin{cases} {\text{HC}}(\sqrt{\lambda_i}|0,\eta) \text{  or}\\ {\text{LU}} (\sqrt{\lambda_i}). \end{cases}
    \end{split}
\end{equation}
The improper Log-uniform distribution has a fatter tail than the Half-Cauchy distribution and is parameter-free. We illustrate both densities in Fig.~\ref{fig: hyperprior}. The Half-Cauchy scaling parameter $\eta>0$ can be adjusted to tune the tradeoff between accuracy and rank-sparsity. Decreasing the magnitude of $\eta$ increases rank-sparsity. Both the Half-Cauchy density function 
\begin{equation}
\label{eq: Half-Cauchy}
{\text{HC}}(x|0,\eta)\propto \left(1+\frac{x^2}{\eta^2}\right)^{-1}
\end{equation}
and the Log-Uniform density function 
\begin{equation}
\label{eq: Log-Uniform}
{\text{LU}} (x)\propto x^{-1}
\end{equation}
place high probability in regions around zero. The parameter $\boldsymbol{\lambda}$ controls the prior variance of the tensor factors in $\boldsymbol\Phi$, all of which have prior mean zero. Therefore the prior density encodes a prior belief that the tensor rank is low, and it encourages structured rank shrinkage. We provide the Bayesian graphical models for each low-rank tensor format in Fig.~\ref{fig: prior description}.

In Fig~\ref{fig: pdf} we demonstrate how our prior induces rank-sparsity in a CP model. Fig~\ref{fig: hyperprior} plots the prior density on the rank parameter $\lambda_j$. Fig.~\ref{fig: marginal prior} shows the corresponding marginal prior on $u_{ij}^{(n)}$. 
The flat tail and sharp peak of the marginal prior induced by the Log-Uniform rank hyper-prior leads to strong shrinkage of small values of $u_{ij}^{(n)}$ towards $0$ but permits medium values to escape the ``gravitational pull" around $0$\cite{carvalho2010horseshoe}. In comparison, the marginal Horseshoe prior induced by the Half-Cauchy hyper-prior exerts a weaker shrinkage effect at small values of $u_{ij}^{(n)}$ but a stronger shrinkage effect on larger values. 

\section{Scalable Parameter Inference}

Now we discuss how to estimate the resulting posterior density~\eqref{eq: posterior}. We develop an efficient tensorized Bayesian inference approach by improving Stochastic Variational Inference (SVI)~\cite{hoffman2013stochastic}. 
We consider SVI~\cite{hoffman2013stochastic} due to its superior computational and memory efficiency over gradient-based MCMC~\cite{neal2012bayesian} and Stein variational gradient descent~\cite{liu2016stein}. However, directly applying SVI to our tensorized training can cause numerical failures. Therefore, we will develop a customized SVI solver with analytical/numerical hybrid parameter update that is suitable for our Bayesian tensorized neural networks. 


\subsection{Review of Stochastic Variational Inference (SVI)}

Let $\boldsymbol{\theta}$ be the parameters to infer and let $q(\boldsymbol{\theta})$ be the approximating distribution to the target posterior distribution 
\[
p(\boldsymbol{\theta}|\mathcal{D})\propto p(\mathcal{D}|\boldsymbol{\theta})p(\boldsymbol{\theta}).
\]
SVI~\cite{hoffman2013stochastic}  solves an optimization problem where the loss function is the KL divergence and the goal is to find the best approximating density $q^\star$ among a parameterized class of densities $\mathcal{P}$:
\begin{equation}
\label{eq: basic svi}
\begin{split}
    q^{\star}(\boldsymbol{\theta}) &= \argmin_{q(\boldsymbol{\theta})\in \mathcal{P}} \text{KL}\left( q(\boldsymbol{\theta})||p(\boldsymbol{\theta}|\mathcal{D})\right),\;\;
    \text{KL}\left( q(\boldsymbol{\theta})||p(\boldsymbol{\theta}|\mathcal{D})\right) = \mathbb{E}_{q(\boldsymbol{\theta})}\left[\log \frac{q(\boldsymbol{\theta})}{p(\boldsymbol{\theta}|\mathcal{D})}\right].
\end{split}
\end{equation}
The KL divergence can be re-written as
\begin{equation}
\label{eq: svi kl}
\begin{split}
    \text{KL}\left( q(\boldsymbol{\theta})||p(\boldsymbol{\theta}|\mathcal{D})\right)       &= \mathbb{E}_{q(\boldsymbol{\theta})}\left[\log {q(\boldsymbol{\theta})}-\log p(\mathcal{D}|\boldsymbol{\theta})-\log p(\boldsymbol{\theta}) \right]+\text{const.}\\    
    &=  -\mathbb{E}_{ q(\boldsymbol{\theta})}\left[\log {p(\mathcal{D}|\boldsymbol{\theta})}\right]+\text{KL}\left( q(\boldsymbol{\theta})||p(\boldsymbol{\theta})\right) +\text{const.}
\end{split}
\end{equation}
This is a combination of the log-likelihood (model fit) and the divergence from the approximate posterior to the prior (low-rank). To approximate the log-likelihood one samples from the variational distribution $q$. The KL-divergence is either approximated via sampling or evaluated in a closed form. The form in Equation (\ref{eq: svi kl}) requires the evaluation of the full-data model likelihood. If the data is large the full-data likelihood $p(\mathcal{D}|\boldsymbol{\theta})$ is intractable, so we approximate the likelihood by subsampling a minibatch $\mathcal{M}\subset \mathcal{D}$.

\subsection{Challenges in Training Bayesian Tensorized Neural Networks} 

Now we explain the challenges of directly applying SVI to train our Bayesian tensorized neural network model. As an example, we focus our notation on the CP-format one-layer model with parameters 
\begin{equation}
\label{eq: SVI CP description}
\boldsymbol{\theta}= \{\boldsymbol\Phi, \boldsymbol{\Lambda}\}=\{\{\fm{n}\}_{n=1}^d,\boldsymbol{\lambda}\}.
\end{equation} 
The extension to other tensor formats and to multiple layers is trivial. For notational convenience we omit the description of the bias term $\mat{b}$ since it is assigned a Normal variational posterior and follows the standard update rules specified in \cite{blundell2015weight}. 

In variational inference, it is a common practice to simplify a posterior density in order to reduce the computational cost. In our problem setting, we firstly use the mean-field approximation~\cite{jordan1999introduction} to achieve a tractable optimization:
\begin{equation}
    \label{eq: mean field}
    \begin{split}
    q\left(\{\fm{n}\},\boldsymbol{\lambda}\right) &= q_{\bf{U}}\left(\{\fm{n}\}\right)q_{\boldsymbol{\lambda}}\left(\boldsymbol{\lambda}\right).
    \end{split}
\end{equation}
We further model the posterior of the tensor factors with a normal distribution 
\begin{equation}
    \label{eq: factor variational distribution}
    \begin{split}
        q_{\bf{U}}\left(\{\fm{n}\}\right) &= \prod \limits_{n=1}^d q_{\fm{n}}\left(\fm{n}\right),\;\;
        q_{\fm{n}}\left(\mat{U}^{(n)}\right) = \prod_{i,j} \mathcal{N}\left(u^{(n)}_{ij}|\overline{u^{(n)}_{ij}},{\Sigma^{(n)}_{ij}}^{2} \right), 
    \end{split}
\end{equation}
where $\overline{u^{(n)}_{ij}}$ and $\Sigma^{(n)}_{ij}$ are the $(i,j)$-th elements of the unknown posterior mean $\overline{\fm{n}}$ and posterior standard deviation $\mat\Sigma^{(n)}$ to be inferred, respectively.

Now we discuss the challenges in learning the variational posterior distribution. We modify Eq. (\ref{eq: svi kl}) to obtain our objective function:
\begin{equation}
\label{eq: low-rank objective}
\begin{split}
    \mathcal{L}\left(q\right)=-\mathbb{E}_{q\left(\{\fm{n}\},\boldsymbol{\lambda}\right)} & \log \: { p(\mathcal{D}|\{\fm{n}\})}+\text{KL}\left( q\left(\{\fm{n}\},\boldsymbol{\lambda}\right)|| p(\{\fm{n}\},\boldsymbol{\lambda})\right).
\end{split}
\end{equation}
Due to the nonlinear tensorized forward model, we need to employ gradient-based iterations in SVI to update the tensor factor parameters. The expected log-likelihood in Equation (\ref{eq: low-rank objective}) must be approximated by sampling the variational distribution $q$. 
The first standard approach is to select a variational distribution $q\left(\{\fm{n}\},\boldsymbol{\lambda}\right)$ for which the KL divergence in Equation (\ref{eq: low-rank objective}) can be obtained in a closed form. The second standard approach is to approximate the KL divergence term by sampling from the variational posterior. In practice, two challenges prevent us from applying these standard SVI approaches:  
\begin{itemize}
    \item {\bf Challenge 1: Closed-form objectives require multiple training runs:} Variational distributions $q$ that permit a closed-form approximation of the KL divergence require additional hyperparameters. Existing distributions that enable a closed-form KL divergence require a hierarchical Bayesian parameterization of the rank parameter $\boldsymbol\lambda$ \cite{wand2011mean,ghosh2019model}, requiring up to five additional hyperparameters for the new random variables \cite{wand2011mean}. Additional hyperparameters would require additional tuning runs and remove the benefits of one-shot tensorized training. Therefore, we avoid this option.
    \item {\bf Challenge 2: Sampling-based approximation increases gradient variance:} Sampling-based approximation of the KL divergence leads to gradient instability during rank shrinkage. The gradient variance with respect to the low-rank tensor factor parameters is proportional to the variance of $1/\boldsymbol\lambda$, and it may explode during rank-shrinkage as $\boldsymbol\lambda$ approaches $ 0$, so sampling $\boldsymbol\lambda$ is not feasible. 
\end{itemize}

\begin{figure}
    \centering
    \includegraphics[width=0.5\textwidth]{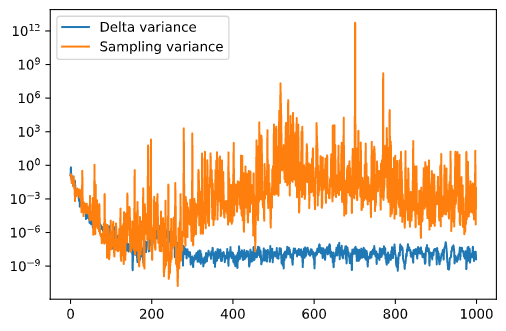}
    \caption{The gradient variance of a single low-rank tensor factor parameter. Sampling the rank parameter $\boldsymbol\lambda$ leads to high-variance gradients, while our proposed delta approximation of hyper parameters reduces the gradient variance significantly (see Section~\ref{subsec: delta posteiror} and Section~\ref{label:subsec gradient update}).}
    \label{fig: gradient variance}
    \vspace{-20pt}
\end{figure}

We provide more details about the second challenge. We consider the gradient of the objective function in Eq. \ref{eq: low-rank objective} w.r.t. the parameters $\overline{u^{(n)}_{ij}}$ and $\Sigma^{(n)}_{ij}$. First we observe that 
\begin{equation}
\label{eq: KL proportionality}
\begin{split}
     \text{KL}\left(\mathcal{N}\left(u^{(n)}_{ij}|\overline{u^{(n)}_{ij}},{\Sigma^{(n)}_{ij}}^2 \right)|| \mathcal{N}\left(u^{(n)}_{ij}|0,{\lambda}_j \right)\right) &\propto \frac{\overline{u^{(n)}_{ij}}^2+{{\Sigma}^{(n)}_{ij}}^2}{{\lambda}_j}.
\end{split}
\end{equation}
 Let $\phi$ represent either parameter of $\left\{\overline{u^{(n)}_{ij}},\Sigma^{(n)}_{ij}\right\}$. Then sampling $\boldsymbol\lambda$ yields a gradient variance
\begin{equation}
\label{eq: KL gradient variance}
\small
\begin{split}
     \mathbb{V}\left[\nabla_\phi \text{KL}\left( \mathcal{N}\left(u^{(n)}_{ij}|\overline{u^{(n)}_{ij}},{\Sigma^{(n)}_{ij}}^2 \right)|| \mathcal{N}\left(u^{(n)}_{ij}|0,{\lambda}_j \right)\right)\right]&\propto \mathbb{V}\left[\frac{1}{{\lambda}_j}\right]. 
\end{split} 
\end{equation}
The goal of our low-rank prior is to shrink many ${\lambda}_j$'s to $0$ in the training process. If the distribution of ${\lambda}_j$ is non-degenerate, even small uncertainties in the value of ${\lambda}_j$ will lead to large variance in Equation \eqref{eq: KL gradient variance} as the posterior probability of ${\lambda}_j$ concentrates around $0$ . As a result, a rank shrinkage can cause high-variance gradients which in turn may increase the magnitude of factor matrix parameters, as shown in Fig. \ref{fig: gradient variance}. 

\subsection{Simplified Posterior for Rank-Controlling Hyper Parameters}
\label{subsec: delta posteiror}

To avoid gradient variance explosion, we propose a deterministic approximation to the hyper parameter $\boldsymbol{\lambda}$: 
\begin{equation}
    \label{eq: variational rank distribution}
        q_{\boldsymbol{\lambda}}(\boldsymbol{\lambda}) = \delta_{\overline{\boldsymbol{\lambda}}}(\boldsymbol{\lambda})
\end{equation}
where $\delta$ is a Delta function and $\overline{\boldsymbol{\lambda}}$ is the posterior mean of $\boldsymbol{\lambda}$. This delta approximation was used for empirical partially Bayes estimation in~\cite{nakajima2014analysis}. This approximation admits {\it closed-form updates} to the following sub-problem when the factor matrices are fixed: 
\begin{equation}
    \label{eq: KL divergence lambda}
     \argmin_{\overline{{\lambda}}_k} \text{KL}\left( q\left(\{\fm{n}\},\boldsymbol{\lambda}\right)|| p(\{\fm{n}\},\boldsymbol{\lambda})\right).
\end{equation}

We provide the closed-form analytical updates for $\overline{\lambda}_k$ under each choice of prior in CP format and give the details in Appendix A. The results associated with other tensor formats can be obtained similarly. For the Log-Uniform prior
\begin{equation}
\small
    \label{eq: lambda update appendix}
    \begin{split}
            \overline{\lambda}_k^\star&\leftarrow \frac{M}{D+1}.\\
\end{split}
\end{equation}
Here we have used the notations
\begin{equation}
\label{eq: notation for update}
\begin{split}
D&=\sum_{n} I_n, \quad
M=\sum_{1\leq n \leq d}\sum_{1\leq i \leq I_n} \overline{u^{(n)}_{ik}}^2+{{\Sigma}^{(n)}_{ik}}^2.\\ 
\end{split}    
\end{equation}
The number of entries controlled by ${\lambda}_j$ is $D$, and $M$ is their combined magnitude and variance. In the case of the Half-Cauchy prior with scale parameter $\eta$, the update is
\begin{equation}
    \label{eq: half cauchy rank parameter update}
        \overline{\lambda}_k^\star \leftarrow\frac{M-\eta^2D+\sqrt{M^2+(2D+8)\eta^2 M +\eta^4 D^2}}{2D+2}.
\end{equation}
For the Half-Cauchy hyperprior, decreasing the magnitude of the scale parameter $\eta$ decreases the magnitude of the update of $\overline{\lambda}_k^\star$, thereby increasing rank-sparsity. 

\subsection{Analytical/Numerical Hybrid Parameter Update in SVI} 
\label{label:subsec gradient update}
With the proposed Delta posterior approximation for $\boldsymbol{\lambda}$, now we can train our tensorized neural network training with an analytical/numerical hybrid parameter update rule in SVI. Specifically, in every iteration of SVI, we use a gradient-based half step to update the tensor factors in $\boldsymbol{\Phi}$ and  closed-form half step to the hyper parameters $\overline{\boldsymbol{\lambda}}$. We apply the reparametrization trick
\begin{equation}
    \label{eq: reparametrization trick}
    u^{(n)}_{ij} = \overline{u^{(n)}_{ij}}+z\Sigma^{(n)}_{ij},\hspace{0.2in} z\sim \mathcal{N}(0,1)
\end{equation}
to sample from the tensor factor distributions.
\begin{itemize}
    \item {\bf Half Step 1: Gradient Update for tensor factors:} We sample the low-rank tensor factors $\boldsymbol\Phi$ and update all parameters of the tensor factor variational distributions using gradient descent on the loss $\mathcal{L}\left(q\right)$ of Eq. \eqref{eq: low-rank objective} with a learning rate $\alpha$:
    \begin{equation}
    \label{eq: gradient update}
\boldsymbol\Phi\leftarrow \boldsymbol\Phi+ \alpha \nabla_{\boldsymbol\Phi} \mathcal{L}\left(q\right).
\end{equation}
In the the CP model, the gradients for the posterior variance and mean of the factor matrices are given by
\begin{equation}
    \label{eq: cp gradient update}
    \begin{split}
\nabla_{\Sigma^{(n)}_{ij}} \mathcal{L}\left(q\right)&= -z\nabla_{u_{ij}}\log \: { p(\mathcal{D}|\{\fm{n}\})}-\frac{1}{\Sigma^{(n)}_{ij}}+\frac{\Sigma^{(n)}_{ij}}{\overline\lambda_j}\\ 
\nabla_{\overline{u^{(n)}_{ij}}}\mathcal{L}\left(q\right)&= -\nabla_{u_{ij}}\log \: { p(\mathcal{D}|\{\fm{n}\})}+\frac{\overline{u^{(n)}_{ij}}}{\overline\lambda_j}.
    \end{split}
\end{equation}
Note that $z$ is the random variable sampled during the forward pass due to the reparameterization in Eq. \eqref{eq: reparametrization trick} and the gradients with respect the log-likelihood are computed using standard automatic differentiation. We describe the gradients for the other three tensor formats in Appendix B.
    \item {\bf Half Step 2: Incremental closed-form update for $\overline{\boldsymbol{\lambda}}$:} We analytically update the rank-controlling parameters $\boldsymbol{\lambda}$ based on the results in \eqref{eq: lambda update appendix} and \eqref{eq: half cauchy rank parameter update}.  We found empirically that incremental updates, rather than direct assignment of the results from (\ref{eq: lambda update appendix}) or (\ref{eq: half cauchy rank parameter update}), led to better performance. Therefore we adopt an incremental update strategy with learning rate $\gamma$ for the rank parameter updates:
\begin{equation}
    \label{eq: rank parameter general update}
    \overline{\lambda}_k \leftarrow \gamma \overline{\lambda}_k^\star+(1-\gamma)\overline{\lambda}_k.
\end{equation}
\end{itemize}

As shown in Figure~\ref{fig: gradient variance}, this proposed hybrid parameter update can greatly reduce the gradient variance of tensor factors.

\subsection{Algorithm Flow and Implementation Issues} The full description of our end-to-end tensorized training with rank determination is shown in Alg.~\ref{alg: training}. We iteratively repeat the hybrid parameter updates for a predetermined number of epochs $m$. In the following, we discuss some important implementation issues.

\begin{algorithm}[t]
\caption{SVI-Based Tensorized Training with Rank Determination}
\begin{algorithmic}
\label{alg: training}
\STATE {\bf Input: }Factor learning rate $\alpha$, EM stepsize $\gamma$, rank cutoff $\epsilon$, warmup epochs $e_w$, total epochs $m$, tuning epochs $t$
\FOR{Epoch $e$ in $[1,\dots, m]$}
\STATE Assign $\beta$ according to Equation (\ref{eq: kl re-weighting}).
    \FOR{each batch $\mathcal{B}\subset \mathcal{D}$}
        \STATE Update the low-rank factor distribution variational parameters as in Half Step 1, Equation \eqref{eq: gradient update}.
        \STATE Update the rank-control hyper-parameters as in Half Step 2, Equation \eqref{eq: rank parameter general update}.
    \ENDFOR
\ENDFOR
\STATE Prune tensor ranks as described in Equation \eqref{eq: rank pruning}.
\end{algorithmic}
\end{algorithm}

\paragraph{Warmup Schedule} A general challenge in Bayesian tensor computation is that poor initializations can lead to excessive rank shrinkage and trivial rank-zero solutions. In linear tensor problems such as tensor completion the SVD is used to generate high-quality initializations \cite{zhao2015bayesian,zhao2016bayesian}. For nonlinear tensorized neural networks we randomly initialize the factor matrices so the predictive accuracy is low and the KL divergence to the prior may dominate the local loss landscape around the initialization point. To avoid trivial rank-zero local optima early in the training process, we incrementally re-weight the KL divergence from the variational approximation to the prior during the training process. Let $e_w$ be the number of warmup training epochs and $e$ be the current epoch. We re-weight the KL divergence from the variational approximation to the prior by a factor $\beta$ defined by
\begin{equation}
    \label{eq: kl re-weighting}
            \beta = \min\left(1,\frac{e}{e_w}\right),
\end{equation}
and update the loss from Eq. (\ref{eq: low-rank objective}) accordingly: 
\begin{equation}
\label{eq: annealed low-rank objective}
\small
\begin{split}
    \mathcal{L}\left(q\right)=-\log &  \mathbb{E}_{q\left(\{\fm{n}\},\boldsymbol{\lambda}\right)} {p(\mathcal{D}|\{\fm{n}\})}+\beta\text{KL}\left( q\left(\{\fm{n}\},\boldsymbol{\lambda}\right)|| p(\{\fm{n}\},\boldsymbol{\lambda})\right).
\end{split}
\end{equation}
Gradually increasing the weight of the KL divergence to the prior avoids early local optima in which all ranks shrink to zero. We have found empirically that $e_w=m/2$ is a good choice for the number of warmup steps. 

\paragraph{ Rank Pruning} After we run our Bayesian solver we truncate the ranks with variance $\overline{\lambda}_k$ below a pre-specified threshold $\epsilon$. For example, for the CP format if $\overline{\lambda}_k<\epsilon$ we assign
\begin{equation}
    \label{eq: rank pruning}
    \begin{split}
        \overline{u^{(n)}_{ik}}&\leftarrow 0 \:\text{and} \: \Sigma^{(n)}_{ik} \leftarrow 0 \: \text{ for }1\leq n \leq d,1\leq i \leq I_n. 
    \end{split}
\end{equation}
The associated $k$-th column of $\mat{U}^{(n)}$ is removed, leading to a rank shrinkage and automatic model parameter reduction.


\begin{table}[t]
{\footnotesize
\caption{Summary of different training methods. \label{table: comparison methods}}
\begin{center}
\begin{tabular}{|c|c |c |c|}
\hline 
Method & memory cost of training & \# training runs & model size for inference  \\ \thickhline
Baseline & high & 1 & huge\\ \hline
FR~\cite{novikov2015tensorizing} & low & many & small\\ \hline
TC-MR ~\cite{lebedev2014speeding} & high & 1 & small \\ \hline
TC-OR ~\cite{lebedev2014speeding} &   high & 1 & small \\ \hline
ARD-LU (Proposed) & low & 1 & small \\\hline
ARD-HC (Proposed) & low & 1 & small \\ \hline
\end{tabular}
\end{center}
}
\end{table}

\section{Experiments}
\label{sec:experiments}

We demonstrate the applications of our rank-adaptive tensorized end-to-end training method on several neural network models. Our method trains a Bayesian neural network, therefore we report the predictive accuracy of the posterior mean. In order to compare the performance, we implement the following methods in our experiments: 
\begin{itemize}
    \item {\bf Baseline:} a standard training method, where model parameters are uncompressed. 
    
        \item {\bf TC-MR}~\cite{lebedev2014speeding,kim2015compression}: train and then compress with maximum ranks. We train a uncompressed neural network with the ``baseline" method, followed by a tensor decomposition and fine-tuning. For the DLRM model we fine-tune for one epoch. In all other experiments we fine-tune for 20 epochs. This approach requires that the user select the compression rank. Here we use the maximum rank used in our Bayesian model. This approach has been studied for computer vision tasks using the CP decomposition in \cite{lebedev2014speeding} and the Tucker decomposition in \cite{kim2015compression,gusak2019musco}. We compare against the algorithms of \cite{lebedev2014speeding,kim2015compression}, but on different architectures.
        \item {\bf TC-OR}: train and then compress with oracle rank ($r$ in CP or $\mat{r}=[r_1, r_2, \cdots, r_d]$ for other formats). This method follows the same procedures of TC-MR~\cite{lebedev2014speeding,kim2015compression}, except that it uses the ``oracle rank" discovered by our proposed rank determination method. In practice this ``TC-OR" method would require a combinatorial rank search over a high-dimensional discrete space to discover the same rank as our method. 
    \item {\bf FR}: Fixed-rank tensorized training. We implement tensorized training~\cite{novikov2015tensorizing,garipov2016ultimate,calvi2019compression,hrinchuk2020tensorized} with a tensor rank fixed {\it a priori}. Determining the tensor ranks is challenging in this approach. In our experiments we reuse the well-tuned parameters from previous literature. The CNN experiment and architecture in the supplemental material is taken from \cite{garipov2016ultimate}. The NLP and DLRM experiment architectures are taken from \cite{hrinchuk2020tensorized}.  
   \item {\bf ARD-LU:} the first version of our proposed tensorized training method with automatic rank determination. We use the log-uniform prior in \eqref{eq: Log-Uniform} for the rank-control hyper-parameters.  All tensor factors are initialized with a maximum rank ($R$ for CP and $\mat{R}=[R_, \cdots, R_d]$ for other formats), and the actual ranks ($r$ for CP and $\mat{r}=[r_1,\cdots r_d]$ for other formats) are automatically determined by our training process. To compare our method with FR, we set the maximum rank to the rank used in FR.
    \item {\bf ARD-HC:} the second version of our proposed training method using the half-Cauchy prior~\eqref{eq: Half-Cauchy} for the rank-control hyper parameters. 
    
\end{itemize}
As shown in Table~\ref{table: comparison methods} our proposed methods enjoy all of the listed advantages compared with other methods. The proposed automatic tensor rank determination avoids the expensive multiple training runs in FR, and it also results in the (almost) smallest models for inference. We consider four low-rank tensor formats for each tensorized method. Therefore, our experiments involve the implementation of 21 specific methods in total (20 tensorized implementations plus one baseline method). For all experiments we list the full tensor dimension and rank settings in the supplement. For all experiments we set the rank parameter learning rate $\gamma=0.9$.


\begin{remark}
In our Bayesian training, every tensorized model parameter is equipped with two training variables (i.e., posterior mean and variance). Therefore the number of training variables is $2\times$ that of the tensorized model parameter numbers. This parameter overhead in Bayesian training brings in the capability of uncertainty quantification in output prediction, which is important for safety-critical applications. Our Bayesian model also allows a point-wise maximum-a-posterior (MAP) training. In MAP training, the only additional parameters required are the rank-control parameters so the number of training variables is only slightly larger than the number of training variables in fixed-rank tensorized training.  
\end{remark}

\subsection{Synthetic Example for Rank Determination}

\begin{figure}[t]
\centering
\begin{subfigure}{0.24\textwidth}{%
\includegraphics[width=\textwidth]{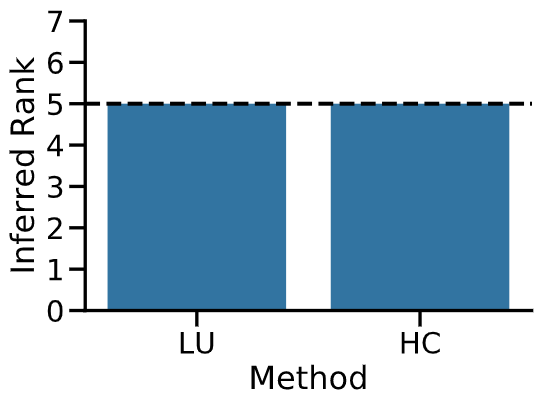}}%
\caption{CP format \label{fig:cp synthetic ranks}}
\end{subfigure}
\begin{subfigure}{0.24\textwidth}{%
\includegraphics[width=\textwidth]{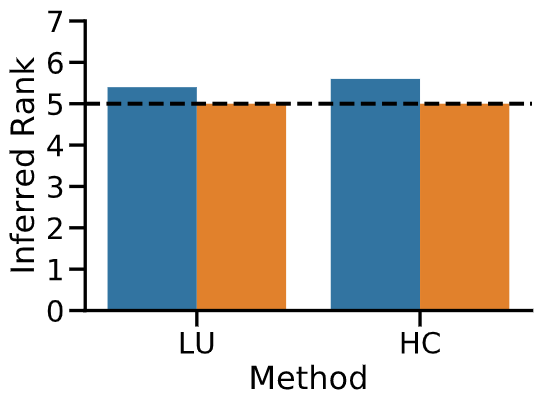}}%
\caption{TT format \label{fig:tt synthetic ranks}}
\end{subfigure}
\begin{subfigure}{0.24\textwidth}{%
\includegraphics[width=\textwidth]{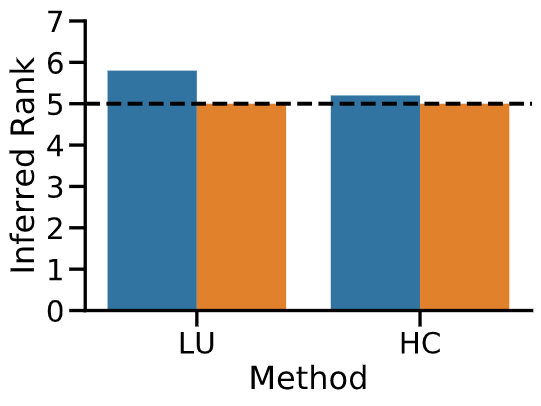}}%
\caption{TTM format \label{fig:ttm synthetic ranks}}
\end{subfigure}
\begin{subfigure}{0.24\textwidth}{%
\includegraphics[width=\textwidth]{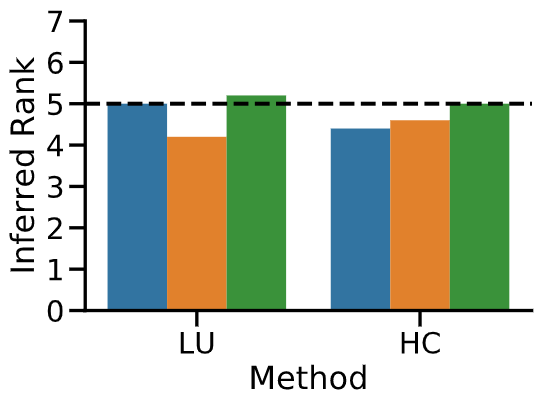}}%
\caption{Tucker format \label{fig:tucker synthetic ranks}}
\end{subfigure}
\caption{The inferred ranks for a synthetic example. The true rank (dashed lines) is $5$ and maximum rank is set to $10$. The inferred ranks of different modes are given by colored bars. \label{fig: synthetic inferred ranks}}
\vspace{-15pt}
\end{figure}

First we test the ability of our proposed method to infer the tensor rank of model parameters in a neural network. For each tensor format we construct a synthetic version of the MNIST dataset using a one-layer tensorized neural network (equivalent to tensorized logistic regression). The tensorized layer is fully connected and the fixed tensor rank is five for each tensor format: 5 for CP, [5, 5, 5] for Tucker and [1, 5, 5, 1] for TT/TTM. We use the rank-5 model to generate synthetic labels for the MNIST images. Then we train a set of low-rank tensorized models with a maximum rank of 10 on the synthetic dataset. For the CP, tensor-train, and Tucker formats we reshape the weight matrix $\mat{W}\in \mathbb{R}^{784\times 10}$ into a tensor of shape size $[28,28,10]$ (i.e., an order-3 tensor of size $28\times28\times 10$). For the tensor-train matrix format we use the dimensions $[4,7,4],[7,2,5]$.

We plot the mean inferred ranks for our log-uniform (LU) and half-cauchy (HC) priors in Fig.~\ref{fig: synthetic inferred ranks}. The actual CP rank is exactly recovered in our model. The inferred ranks of Tucker, TT and TTM are close to but not equal to the exact values, because tensor ranks are not unique, which is a fundamental difference between matrices and tensors.

\subsection{MNIST}

\begin{figure} [t]
\centering
\begin{subfigure}{0.22\textwidth}{%
\includegraphics[width=\textwidth]{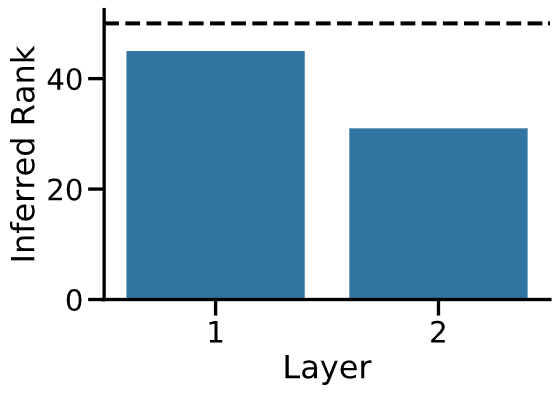}}%
\caption{CP format \label{fig:cp inferred ranks}}
\end{subfigure}
\begin{subfigure}{0.22\textwidth}{%
\includegraphics[width=\textwidth]{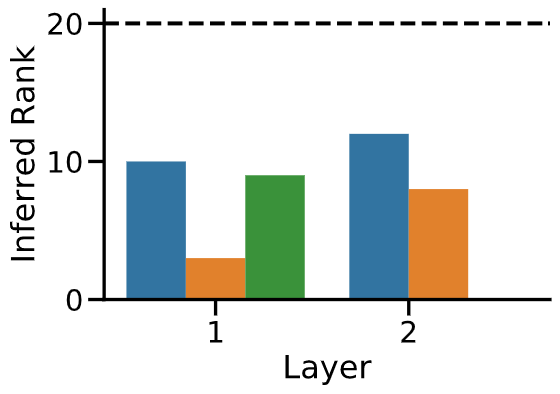}}%
\caption{TT format \label{fig:tt inferred ranks}}
\end{subfigure}
\begin{subfigure}{0.22\textwidth}{%
\includegraphics[width=\textwidth]{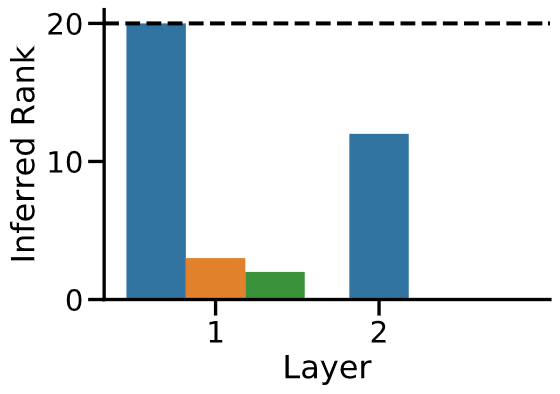}}%
\caption{TTM format \label{fig:ttm inferred ranks}}
\end{subfigure}
\begin{subfigure}{0.22\textwidth}{%
\includegraphics[width=\textwidth]{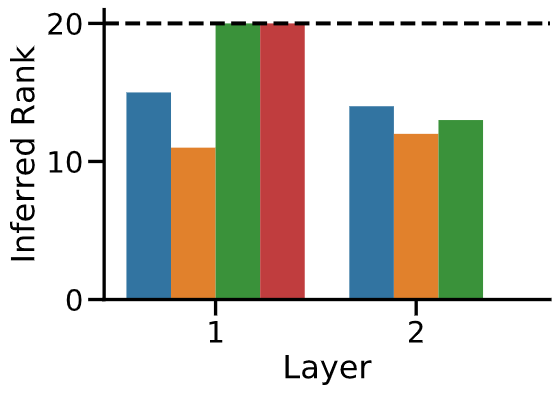}}%
\caption{Tucker format \label{fig:tucker inferred ranks}}
\end{subfigure}
\caption{Inferred ranks for one run of the MNIST experiment using a log-uniform prior. The maximum rank is given by a dashed black line. The inferred ranks are given by colored bars. \label{fig: mnist inferred ranks}}
\vspace{-10pt}
\end{figure}

Next we test a neural network with two fully connected (FC) layers on the MNIST dataset with images of size $28\times 28$. The first FC layer is size $784\times 512$ and has a ReLU activation function. The second FC layer is size $512\times 10$ with a softmax activation function. Exact tensor dimensions are given in Table 2 of Appendix D. 
In all cases our automatic rank determination can achieve the highest compression ratio in training. Our proposed automatic rank determination both improves accuracy and reduces parameter number in all tensor formats except the TT format which has slight accuracy loss but the highest compression ratio. We hypothesize that the automatic rank reduction can reduce over-fitting on the simple MNIST task. The TTM format is best-suited to fully connected layers, achieving the second-highest compression ratios and the second-best accuracy. In Fig.~\ref{fig: mnist inferred ranks} we plot the rank determination output of a single training run using our log-uniform prior. We note that our algorithm discovers the actual ranks that are nearly impossible to determine via hand-tuning or combinatorial search (for example [1,20,3,2,1] in the TTM model from a maximum rank of [1,20,20,20,1], which may require up to 16,000 searches).

\begin{figure} [t]
\centering
\includegraphics[width=\textwidth]{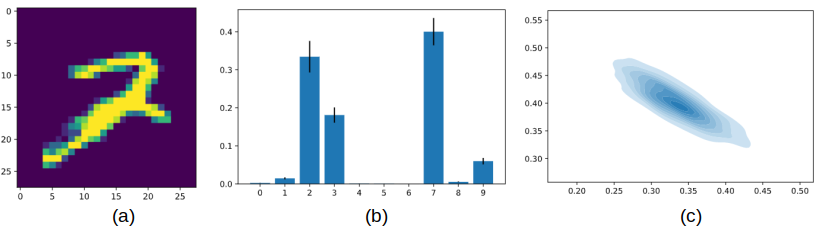}
\caption{(a) A challenging MNIST image with true label ``2". (b) Mean and standard deviation of the CP ARD-LU model softmax outputs. (c) Marginal predictive density of the two most likely labels ``2" (x-axis) and ``7" (y-axis).\label{fig: mnist challenge}}
\vspace{-10pt}
\end{figure}

With the obtained Bayesian solution, we can quantify the uncertainty of our model as a by-product. Popular metrics for uncertainty measures include negative log-likelihood, expected calibration error, which measures model over-/under-confidence, and out-of-distribution input detection~\cite{lakshminarayanan2017simple}. In Fig.~\ref{fig: mnist challenge}, we show the classification uncertainty of an image that is hard to recognize in practice. With the CP tensorized model trained from ARD-LU, we plot the mean and variance of the predicted softmax outputs in Fig.~\ref{fig: mnist challenge} (b). This plot clearly shows that this image looks like ``2", ``3" or ``7", with the highest probability of being classified as ``7".  Fig.~\ref{fig: mnist challenge} (c) further plots the marginal predictive density of the two most likely labels ``2" and ``7".

\begin{table}[t]
\begin{center}
 \begin{threeparttable}
{\footnotesize
\caption{Training results of the MNIST example. \label{table: mnist results}}
\begin{tabular}{|c|c|c|c|c|}
\hline
Tensor Format & Model & Training Parameter $\#$
& Final Parameter $\#$ & Accuracy\\ \thickhline
& Baseline &  407,050  & 407,050 & 98.09  \\ \hline
& FR  &  8,622 ($47.2\times$) & 8,622 ($47.2\times$) & 97.52 \\
& TC-MR ~\cite{lebedev2014speeding}& 407,050 ($1\times$) & 8,622 ($47.2\times$) &  97.32 \\
CP& TC-OR ~\cite{lebedev2014speeding} & 407,050 ($1\times$) &7,175 ($56.7\times$) &  97.36 \\
& ARD-LU (Proposed) &  17,344 ($23.5\times$)  &7,175 ($56.7\times$)  & 98.06 \\
& ARD-HC (Proposed) &  17,344 ($23.5\times$)  &7,134 ($57.1\times$)  &  97.98  \\ \hline
& FR ~\cite{calvi2019compression} &  171,762 ($2.4\times$)  &171,762 ($2.4\times$)  & 97.93 \\
& TC-MR ~\cite{kim2015compression} & 407,050 ($1\times$) & 171,762 ($2.4\times$)  & 98.00 \\
Tucker & TC-OR ~\cite{kim2015compression} & 407,050 ($1\times$) &100,758 ($4.0\times$)   & 97.91  \\
& ARD-LU (Proposed) & 343,644 $(1.18\times)$ & 100,758 ($4.0\times$)   & 98.30  \\
& ARD-HC (Proposed) & 343,644 $(1.18\times)$ & 91,332 ($4.5\times$)   & 98.30  \\ \hline
& FR ~\cite{novikov2015tensorizing} & 26,562 ($13.9\times$)  &
26,562 ($15.3\times$)    & 97.78 \\
& TC-MR  &407,050 ($1\times$) &  26,562 ($15.3\times$)    & 97.43 \\
TT & TC-OR & 407,050 ($1\times$) & 4,224 ($96.4\times$) &  96.91 \\
& ARD-LU (Proposed) & 53,224 $(7.65\times)$ & 4,224 ($96.4\times$) & 96.28 \\
& ARD-HC (Proposed) & 53,224 $(7.65\times)$ & 4,276 ($95.2\times$) & 97.04 \\ \hline
& FR ~\cite{novikov2015tensorizing} &  29,242 ($13.9\times$) &  29,242 ($13.9\times$) & 98.06 \\ 
& TC-MR  & 407,050 ($1\times$) & 29,242 ($13.9\times$)  & 97.47 \\
TTM & TC-OR  & 407,050 ($1\times$) &6,144 ($66.3\times$) & 96.61 \\
 & ARD-LU (Proposed) & 58,564 $(6.95\times)$ & 6,144 ($66.3\times$)  & 98.24 \\
& ARD-HC (Proposed) & 58,564 $(6.95\times)$ &5,200 ($78.3\times$)  & 98.23 \\ \hline
\end{tabular}}
\begin{tablenotes}
      \small
      \item Note: the training parameters in ARD-LU and ARD-HC include posterior mean and variance, so the training parameter number is $2\times$ of that in FR. The results of FR rely on manual rank tuning in contrast to our automatic rank determination procedure.
    \end{tablenotes}
 \end{threeparttable}
\end{center}
\vspace{-10pt}
\end{table}

\subsection{Embedding Table for Natural Language Processing (NLP)}

\begin{table}[t]
\begin{center}
  \begin{threeparttable}
\caption{Training results on the NLP embedding table. \label{table: embedding results} }

{\footnotesize
\begin{tabular}{|c|c|c|c|c|}
\hline
Tensor Type & Model & Training & Final model  & Accuracy\\ 
& &Parameter $\#$ & Parameter $\#$&\\ \thickhline
& Baseline &  6,400,000  & 6,400,000 & 88.34 \\ \hline
& FR  & 8,276 (774$\times$) & 8,276 (774$\times$) & 87.44 \\
&TC-MR  &  6,400,000 $(1\times)$ & 8,276 (774$\times$) & 74.46 \\
CP &TC-OR  & 6,400,000 $(1\times)$&6,138 (1024$\times$) & 73.21\\
&ARD-LU (Proposed) & 16,602 $(385\times)$ & 6,138 (1024$\times$) & 87.61 \\
&ARD-HC (Proposed) & 16,602 $(385\times)$& 6,476 (998$\times$) & 87.54 \\ \hline
& FR &  78,540 (81$\times$)  &78,540 (81$\times$)  & 87.80 \\
&TC-MR & 6,400,000 $(1\times)$& 78,540 (81$\times$)  & 75.12 \\
Tucker& TC-OR  & 6,400,000 $(1\times)$&61,920 (103$\times$) & 71.97  \\
&ARD-LU (Proposed) & 157,105 $(40\times)$& 61,920 (103$\times$) & 87.79  \\
&ARD-HC (Proposed) & 157,105 $(40\times)$ & 58,120 (110$\times$) & 88.01  \\ \hline
& FR \cite{hrinchuk2020tensorized}&  28,260 (226$\times$)  &28,260 (226$\times$)  & 85.6 \\
&TC-MR  & 6,400,000 $(1\times)$& 28,260 (226$\times$)  & 82.34  \\
TT & TC-OR & 6,400,000 $(1\times)$ &22,982 (278$\times$)  & 71.81 \\
&ARD-LU (Proposed) & 56,640 $(113\times)$ & 22,982 (278$\times$)  & 85.33 \\
&ARD-HC (Proposed) & 56,640 $(113\times)$ & 19,363 (331$\times$)  & 85.82 \\ \hline
& FR ~\cite{hrinchuk2020tensorized} &  22,312 (287$\times$)  &22,312 (287$\times$)  & 88.59 \\
&TC-MR  & 6,400,000 $(1\times)$& 22,312 (287$\times$)  & 83.79 \\
TTM & TC-OR & 6,400,000 $(1\times)$ &15,932 (402$\times$)   & 84.83 \\
& ARD-LU (Proposed) & 44,724 $(143\times)$ & 15,932 (402$\times$)   & 88.93 \\
& ARD-HC (Proposed) &  44,724 $(143\times)$ & 14,275 (448$\times$) & 88.78 \\ \hline
\end{tabular}}
\begin{tablenotes}
      \small
      \item Note: the training parameters in ARD-LU and ARD-HC include posterior mean and variance of each tensorized model parameter. The results of FR rely on manual rank tuning in contrast to our automatic rank determination procedure.
    \end{tablenotes}
 \end{threeparttable}
\end{center}
\end{table}

We continue to validate our algorithm with a sentiment classification task from~\cite{khrulkov2019tensorized}. Like many NLP models, the first layer is a large embedding table. Embedding tables are a promising target for tensor compression because their required input dimension equals the number of unique tokens in the input dataset (i.e. number of vocabulary words, number of users). Tensor decomposition can enforce weight sharing and dramatically reduce the parameter count of these models. Recent work in tensorized neural networks has applied the TTM format to compress large embedding tables with a high ratio~\cite{khrulkov2019tensorized}. We replicate a sentiment classification model on the IMDB dataset from their work. The neural network model consists of an embedding table with dimension $25,000\times 256$, two bidirectional LSTM layers with hidden unit size 128, and a fully-connected layer with 256 hidden units. Following the setting in~\cite{khrulkov2019tensorized} we do not tensorize these layers. Dropout masks are applied to the output of each layer except the last. Exact tensor dimensions are given in Table 3 of Appendix D.

We test all methods on the sentiment classification problem. The tensor dimensions and maximum ranks used to compress the embedding table are given in the supplementary material. The outcomes of our experiments are reported in Table~\ref{table: embedding results}. Compared with all other tensor approaches, our methods (ARD-LU and ARD-HC) have achieved the best compression ratio for all tensor formats at little to no accuracy cost. The TTM format outperforms all other models (including the baseline uncompressed model) in terms of accuracy, though we note that the CP model performs well despite its extremely low parameter number. 

\subsection{Deep Learning Recommendation System (DLRM)}
We continue to use our proposed Bayesian tensorized method to train the benchmark Deep Learning Recommendation Model (DLRM)~\cite{naumov2019deep}. In the DLRM model, embedding tables are used to process categorical features, while continuous features are processed with a bottom multilayer perceptron (MLP). Then, second-order interactions of different features are computed explicitly. The results are processed with a top MLP and fed into a sigmoid function in order to give a probability of a click. The whole model has over 4 billion training variables.

We tensorize the five largest embedding tables to reduce the training variables. Exact tensor dimensions are given in Table 4 of Appendix D. Our experiment results are reported in Table~\ref{table: dlrm results}. Our proposed automatic rank reduction enables parameter reduction at little to no accuracy cost over fixed-rank tensorized training. Our approach outperforms the train-then-compress approach which requires expensive full-model training. Compared with baseline full-size training, our method achieves to up to $27,664\times$ (in TT format) parameter reduction during training with little accuracy loss. Our one-shot training also greatly increases the compression ratio over fixed-rank training at little to no accuracy cost, enabling up to $7\times$ higher compression ratios in the TTM model.

The train-then-compress approach can be expensive for this large-scale problem. Because the trained embedding tables are extremely large, compressing them in Tucker or CP format is computationally expensive and time-consuming. This challenge can be avoided in our end-to-end-training approaches because we do not need to explicitly form the embedding tables. 



\begin{table}[t]
\caption{Training results on the DLRM embedding tables.}
\begin{center}
\begin{threeparttable}
{\footnotesize
\begin{tabular}{|c|c|c|c|c|}
\hline
Tensor Type & Model & Training & Final model  & Accuracy\\ 
& &Parameter $\#$ &Parameter $\#$ & \\ \thickhline
& Baseline & 4,248,739,968 & 4,248,739,968 & 78.75 \\ \hline
& FR  & 1,141,597 (3,721$\times$)  &1,141,597 (3,721$\times$)  & 78.60 \\
&TC-MR  & 4,248,739,968 ($1\times$) & 1,141,597 (3,721$\times$) & 75.41 \\
CP &TC-OR  & 4,248,739,968 ($1\times$) &  563,839 (7,535$\times$) & 74.92 \\
&ARD-LU (Proposed) & 2,284,844 (1860$\times$) &  563,839 (7,535$\times$)   &78.61 \\
&ARD-HC (Proposed) & 2,284,844 (1860$\times$) & 570,685 (7,444$\times$)   & 78.57 \\ \hline
& FR  &   1,131,212 (3,755$\times$)  &1,131,212 (3,755$\times$)  & 78.60 \\
&TC-MR  &  4,248,739,968 ($1\times$) & 1,131,212 (3,755$\times$)  & 78.67 \\
Tucker& TC-OR & 4,248,739,968 ($1\times$) &  436,579 (9,731$\times$) & 78.50\\
&ARD-LU (Proposed) & 2,262,852 (1,877$\times$) & 436,579 (9,731$\times$) & 78.64 \\
&ARD-HC (Proposed) & 2,262,852 (1,877$\times$) & 402,023 (10,568$\times$)  & 78.62  \\ \hline
& FR ~\cite{hrinchuk2020tensorized} & 1,135,752   (3,740$\times$) & 1,135,752   (3,740$\times$)  & 78.68  \\
&TC-MR & 4,248,739,968 ($1\times$) &1,135,752  (3,740$\times$)  & 78.39  \\
TT & TC-OR & 4,248,739,968 ($1\times$) &153,582  (27,664$\times$ ) & 78.45 \\
&ARD-LU (Proposed) & 2,271,864 (1870$\times$) & 153,582  (27,664$\times$)& 78.67  \\
&ARD-HC (Proposed) & 2,271,864 (1870$\times$) & 159,529 (26,633$\times$) & 78.63 \\ \hline
& FR ~\cite{hrinchuk2020tensorized} & 1,130,048 (3759$\times$)& 1,130,048 (3759$\times$)  & 78.73 \\
&TC-MR &  4,248,739,968 ($1\times$) &1,130,048 ($3759\times$)  & 78.43 \\
TTM & TC-OR & 4,248,739,968 ($1\times$) &199,504 ($21,296\times$) & 78.62  \\
& ARD-LU (Proposed) & 2,260,256 (1879$\times$) &199,504 (21,296$\times$) & 78.72 \\
& ARD-HC (Proposed) & 2,260,256 (1879$\times$)&163,976 (25,910$\times$) & 78.73 \\ \hline
\end{tabular}}
\begin{tablenotes}
      \small
      \item Note: the training parameters in ARD-LU and ARD-HC include posterior mean and variance of every tensorized model parameters, so the number of training variables is $2\times$ of that in fixed-rank tensorized training (FR).The results of FR rely on manual rank tuning in contrast to our automatic rank determination procedure.
    \end{tablenotes}
 \end{threeparttable}
\end{center}
\label{table: dlrm results}
\end{table}

\subsection{Impact: On-Device Training and FPGA Acceleration}

Our method can successfully train large end-to-end tensor compressed neural networks and increase the compression ratio during training. End-to-end compressed training has a major impact on edge device training by reducing off-chip memory reads which are an energy and latency bottleneck \cite{sze2017hardware}. In \cite{zhang2021fpga} a preliminary FPGA acceleration of our method demonstrates $123\times$ gains in energy efficiency and $59\times$ speedup on a simple two-layer neural network over non-tensorized training on embedded device CPU. These latency and efficiency gains show how our method enables practical on-device training of compact neural networks from scratch. We envision that the performance improvement of on-device training will be more significant on large-scale neural networks. This method may also be implemented with distributed training or on multiple FPGAs to improve the energy efficiency of training huge models on HPC or data centers.

\section{Conclusion and Future Work}

This work has proposed a variational Bayesian method for one-shot end-to-end training of tensorized neural networks. Our work has addressed the fundamental challenge of automatic rank determination, which is important for training compact neural network models on resource-constrained hardware platforms. The customized stochastic variational inference method developed in this paper enables us to train tensorized neural networks with billions of uncompressed model parameters. Our experiments have demonstrated that the proposed end-to-end tensorized training can reduce the training variables by several orders of magnitude. Our proposed method has outperformed all existing tensor compression methods on the tested benchmarks in terms of both compression ratios and predictive accuracy. 

This work will enable ultra memory- and energy-efficient training of AI models on resource-constraint computing platforms, as demonstrated by our preliminary on-FPGA tensorized training in~\cite{zhang2021fpga}. We will further investigate the theoretical and algorithm/hardware co-design issues in this direction, especially for training large-size neural networks on resource-constraint computing platforms.


%






\bibliographystyle{IEEEtran}
\bibliography{bib}

\appendix

\newpage
This supplementary document provides more details of the paper submission ``Towards Compact Neural Networks via End-to-End Training: A Bayesian Tensor Approach with Automatic Rank Determination" to SIAM J. Math Data Science by Cole Hawkins, Xing Liu and Zheng Zhang . 

\section{Rank Parameter Updates}

Firstly we explain the analytical update rules of $\overline{\lambda}_k$ presented in Section 4.3 of the body text. We derive the closed-form updates to a single rank parameter $\overline{\lambda}_k$ for the CP model. The results associated with other tensor formats can be obtained similarly. Firstly, we re-arrange the KL divergence to the prior to isolate all terms involving the rank parameter $\overline{\lambda}_k$:
\begin{equation}
\small
    \label{eq: appendix lambda update derivation}
    \begin{split}
             &\text{KL}\biggl( q\left(\{\fm{n}\}, \boldsymbol{\lambda}\right)   || p(\{\fm{n}\},\boldsymbol{\lambda})\biggr)\\
             =&\text{KL}\left( q\left(\{\fm{n}\}\right)|| p\left(\{\mat{U}^{(n)}\}\:|\:\boldsymbol{\lambda}\right) \right)+\text{KL}\left( q\left(\boldsymbol{\lambda}\right)|| p\left(\boldsymbol{\lambda}\right) \right)\\
             =&\sum_{1\leq n\leq d}\sum_{1\leq i \leq I_n, 1\leq j \leq R}\text{KL}\left(\mathcal{N}\left(\overline{u^{(n)}_{ij}},{\Sigma^{(n)}_{ij}}^2 \right)||\mathcal{N}\left(0,{\lambda}_j \right) \right)+\sum_{1\leq j\leq R}\text{KL}\left(\delta({\lambda}_j)||p({\lambda}_j) \right)\\
             \propto & \sum_{1\leq n\leq d}\sum_{1\leq i \leq I_n}\text{KL}\left(\mathcal{N}\left(\overline{u^{(n)}_{ik}},{\Sigma^{(n)}_{ik}}^2 \right)||\mathcal{N}\left(0,{\lambda}_k \right) \right)-p({\lambda}_k)\\
            \propto & \sum_{1\leq n\leq d}\sum_{1\leq i \leq I_n}\left(\log\sqrt{{\lambda}_k}+\frac{\overline{u^{(n)}_{ik}}^2+{\Sigma^{(n)}_{ik}}^2}{2{\lambda}_k}\right) -p({\lambda}_k).
    \end{split}
 \end{equation}
 Next, we consider a log-uniform rank prior $p(\overline\lambda_k)$ and take the derivative of the KL divergence with respect to $\overline{\lambda}_k$. This yields
\begin{equation}
\small
    \label{eq: appendix lambda derivate update appendix}
    \begin{split}
            \frac{\partial}{\partial \overline{\lambda}_k}\text{KL}&\left( q\left(\{\fm{n}\},\boldsymbol{\lambda}\right)|| p(\{\fm{n}\},\boldsymbol{\lambda})\right) \propto \sum_{1\leq n\leq d}\sum_{1\leq i \leq I_n}\left(\frac{1}{2\overline{\lambda}_k}-\frac{\overline{u^{(n)}_{ik}}^2+{\Sigma^{(n)}_{ik}}^2}{2\overline{\lambda}_k^2}\right)  +\frac{1}{2\overline{\lambda}_k}.
\end{split}
\end{equation}
 Finally, enforcing the gradient ~(\ref{eq: appendix lambda derivate update appendix}) to be zero yields a closed-form update:
\begin{equation}
\small
    \label{eq: appendix lambda update appendix}
    \begin{split}
            \overline{\lambda}_k^\star&\leftarrow \frac{M}{D+1}.\\
\end{split}
\end{equation}
Here we have used the notations
\begin{equation}
\label{eq: appendix notation for update}
\begin{split}
D&=\sum_{n} I_n, \quad
M=\sum_{1\leq n \leq d}\sum_{1\leq i \leq I_n} \overline{u^{(n)}_{ik}}^2+{{\Sigma}^{(n)}_{ik}}^2.\\ 
\end{split}    
\end{equation}
The number of entries controlled by ${\lambda}_j$ is $D$, and $M$ is their combined magnitude and variance.

In the case of the Half-Cauchy prior with scale parameter $\eta$, the update is
\begin{equation}
    \label{eq: appendix half cauchy rank parameter update}
        \overline{\lambda}_k^\star \leftarrow\frac{M-\eta^2D+\sqrt{M^2+(2D+8)\eta^2 M +\eta^4 D^2}}{2D+2}.
\end{equation}
Decreasing the magnitude of the scale parameter $\eta$ decreases the magnitude of the update of $\overline{\lambda}_k^\star$, thereby increasing rank-sparsity.

\section{Gradient Updates}\\
In Section 4.4 of the body text, we have provided the gradient update rules for CP tensor factors. Here we provide the gradient update for other three tensor formats in our tensorize neural network training.

\noindent{\bf Tucker:} In Tucker format the gradients take a similar form as in the CP format except the rank parameter is dimension-specific. 
\begin{equation}
    \label{eq: tucker gradient update}
    \begin{split}
\nabla_{\Sigma^{(n)}_{ij}} \mathcal{L}\left(q\right)&= -z\nabla_{u_{ij}}\log \: { p(\mathcal{D}|\ten{G},\{\fm{n}\})}-\frac{1}{\Sigma^{(n)}_{ij}}+\frac{\Sigma^{(n)}_{ij}}{\lambda_j^{(n)}}\\ 
\nabla_{\overline{u^{(n)}_{ij}}}\mathcal{L}\left(q\right)&= -\nabla_{u_{ij}}\log \: { p(\mathcal{D}|\ten{G},\{\fm{n}\})}+\frac{\overline{u^{(n)}_{ij}}}{\lambda_j^{(n)}}\\ 
    \end{split}
\end{equation}
The core tensor $\ten{G}$ is updated according to 
\begin{equation}
\small
    \label{eq: tucker core update}
    \begin{split}
\nabla_{{\Sigma_{i_1\dots i_d}}}&\mathcal{L}\left(q\right)\\
&= -z\nabla_{\Sigma_{i_1\dots i_d}}\log \: { p(\mathcal{D}|\ten{G},\{\fm{n}\})}-\frac{1}{\Sigma_{i_1\dots i_d}}+\frac{\Sigma_{i_1\dots i_d}}{\sigma_0^2}\\
\nabla_{\overline{g}_{i_1\dots i_d}}& \mathcal{L}\left(q\right)\\
&= -\nabla_{\overline{g}_{i_1\dots i_d}}\log \: { p(\mathcal{D}|\ten{G},\{\fm{n}\})}+\frac{\overline{g}_{i_1\dots i_d}}{\sigma_0^2}.\\
    \end{split}
\end{equation}

\noindent{\bf Tensor Train:} In TT-format the parameters are re-indexed to accommodate a third dimension in addition to the dimension-specific rank parameters referenced above.
\begin{equation}
    \label{eq: tt gradient update}
    \begin{split}
\nabla_{\Sigma^{(n)}_{ijk}} \mathcal{L}\left(q\right)&= -z\nabla_{g_{ijk}}\log \: { p(\mathcal{D}|\{\ft{n}\})}-\frac{1}{\Sigma^{(n)}_{ijk}}+\frac{\Sigma^{(n)}_{ijk}}{\lambda_k^{(n)}}\\ 
\nabla_{\overline{g^{(n)}_{ijk}}}\mathcal{L}\left(q\right)&= -\nabla_{g_{ijk}}\log \: { p(\mathcal{D}|\{\ft{n}\})}+\frac{\overline{g^{(n)}_{ijk}}}{\lambda_k^{(n)}}\\ 
    \end{split}
\end{equation}

\noindent{\bf Tensor Train Matrix:} Finally, for the TTM format we perform one additional re-indexing.
\begin{equation}
    \label{eq: ttm gradient update}
    \begin{split}
\nabla_{\Sigma^{(n)}_{ijkl}} \mathcal{L}\left(q\right)&= -z\nabla_{g_{ijkl}}\log \: { p(\mathcal{D}|\{\ft{n}\})}-\frac{1}{\Sigma^{(n)}_{ijkl}}+\frac{\Sigma^{(n)}_{ijkl}}{\lambda_l^{(n)}}\\ 
\nabla_{\overline{g^{(n)}_{ijkl}}}\mathcal{L}\left(q\right)&= -\nabla_{g_{ijkl}}\log \: { p(\mathcal{D}|\{\ft{n}\})}+\frac{\overline{g^{(n)}_{ijkl}}}{\lambda_l^{(n)}}\\ 
\end{split}
\end{equation}

\newpage

\section{Additional Example: CIFAR Convolutional Model}

We provide additional experiments on a convolutional neural network taken from~\cite{garipov2016ultimate}. This model consists of six convolutional layers followed by three fully connected layers. We follow~\cite{garipov2016ultimate} and tensorize all layers except the first convolution and the last fully connected layer which together contain a small fraction of the total parameters. As before, we test all four tensor formats with our rank determination approach. The results of our method, the baseline model, and the train-and-then-compress approach are reported in Table ~\ref{table: appendix cnn results}.

We observe that our proposed method (ARD) leads to higher accuracy than the train-and-then-compress approach. Our automatic rank determination achieves parameter reduction with only slight accuracy reduction. The CP and TTM methods outperform Tucker and TT methods for this task in terms of accuracy. Previous studies~\cite{kim2015compression, garipov2016ultimate} have shown that the compression ratio on convolution layers are often much lower than on fully connected layers due to the small size of convolution filters. Nevertheless, our tensorized training with automatic rank determination always achieves the best compression performance. 

\begin{table}[t]
\begin{center}
 \begin{threeparttable}
{\footnotesize
\caption{Training results on the CNN model. \label{table: appendix cnn results}}
\begin{tabular}{|c|c|c|c|c|}
\hline
Tensor Type & Model & Training Parameter $\#$ & Final Parameter $\#$ & Accuracy\\ \thickhline
& Baseline & 13,942,602 & 13,942,602 & 90.36 \\ \hline
&  FR  &  652,748 (21.4$\times$) & 652,748 (21.4$\times$) & 90.13 \\
& TC-MR \cite{lebedev2014speeding} & 13,942,602 (1$\times$) & 9652,748 (21.4$\times$) & 75.80 \\
CP & TC-OR ~\cite{lebedev2014speeding} & 13,942,602 (1$\times$) & 568,412 (24.5$\times$) & 71.29\\
&ARD-LU (Proposed) & 1,308,418 ($10.6\times$) & 568,412 (24.5$\times$) & 90.18 \\
&ARD-HC (Proposed) & 1,308,418 ($10.6\times$) & 593,419 (23.5$\times$) & 90.08 \\ \hline
& FR ~\cite{calvi2019compression} & 653,438  (21.3$\times$) &653,438  (21.3$\times$) & 85.15 \\
& TC-MR ~\cite{kim2015compression} & 13,942,602 (1$\times$) &653,438  (21.3$\times$)  & 85.36  \\
Tucker & TC-OR ~\cite{kim2015compression} & 13,942,602 (1$\times$) &606,201 (23.0$\times$) & 84.86 \\
& ARD-LU (Proposed) & 1,307,591 (10.7$\times$) & 606,201 (23.0$\times$) & 85.41 \\
& ARD-HC (Proposed) & 1,307,591 (10.7$\times$)& 589,092 (23.7$\times$) & 85.86  \\ \hline
&  FR ~\cite{novikov2015tensorizing} & 649,328 (21.5$\times$)  & 649,328 (21.5$\times$)  & 87.31 \\
& TC-MR  & 13,942,602 (1$\times$) & 649,328 (21.5$\times$) & 86.02 \\
TT & TC-OR &13,942,602 (1$\times$) & 376,123 (37.1$\times$)  & 85.42 \\
& ARD-LU (Proposed) & 1,299,106 ($10.7\times$) & 376,123 (37.1$\times$)  & 86.68 \\
& ARD-HC (Proposed) & 1,299,106 ($10.7\times$) & 521,096 (26.8$\times$)  & 85.92 \\ \hline
& FR ~\cite{novikov2015tensorizing} & 641,898 (21.7$\times$)  &641,898 (21.7$\times$)  & 90.04 \\
& TC-MR  & 13,942,602 (1$\times$) & 641,898 (21.7$\times$)  & 81.88  \\
TTM & TC-OR & 13,942,602 (1$\times$) &598,693 (22.3$\times$)   & 80.49 \\
& ARD-LU (Proposed) & 1,284,586 ($10.9\times$) & 598,693 (22.3$\times$)   & 90.09  \\
& ARD-HC (Proposed) & 1,284,586 ($10.9\times$) & 579,217 (24.1$\times$) & 90.02 \\ \hline
\end{tabular}}
\begin{tablenotes}
      \small
      \item Note: the training parameters in ARD-LU and ARD-HC include posterior mean and variance of every tensorized model parameters, so the number of training variables is $2\times$ of that in fixed-rank tensorized training (FR).
    \end{tablenotes}
 \end{threeparttable}
\end{center}
\end{table}

\section{Tensorization Settings} Finally, we provide details about the tensorized setting of the MNIST example (Section 5.2), NLP example (Section 5.3) and DLRM example (Section 5.4). 

\begin{table}[t]
{\footnotesize
\caption{Tensorization settings for the MNIST example. \label{table: appendix mnist design}}
\begin{center}
\begin{tabular}{|c|c|c|c|}
\hline
Model & Layer 1 Dimensions  & Layer 2 Dimensions & Max Rank  \\ \thickhline
Baseline &  $784\times512$  & $512\times 10$ & NA\\
CP & [28,28,16,32] & [32,16,10] & 50 \\
Tucker &  [28,28,16,32]  & [32,16,10] & 20\\
TT &  [28,28,16,32]  & [32,16,10] & 20 \\
TTM &  [4,7,4,7],[4,4,8,4]  & [32,16],[2,5] & 20 \\ \hline
\end{tabular}
\end{center}
}
\end{table}

\begin{table}[t]
{\footnotesize
\caption{Tensorization settings for the NLP embedding table. \label{table: appendix embedding design}}
\begin{center}
\begin{tabular}{|c|c|c|c|}
\hline
Model & Embedding Dimensions   & Max Rank  \\ \thickhline
Baseline &  $25,000\times 256$   & NA\\
CP & [5,8,25,25,4,8,8] & 50 \\
Tucker &  [25,25,40,16,16]   & 5\\
TT &  [5,8,25,25,4,8,8]   & 20 \\
TTM &  [5,5,5,5,6,8],[2,2,2,2,4,4] & 20 \\ \hline
\end{tabular}
\end{center}
}
\end{table}

\begin{table}[t]
{\footnotesize
\caption{Tensorization settings for the DLRM embedding tables.}
\label{table: appendix dlrm embedding full design}
\begin{center}
\begin{tabular}{|c|c|c|c|c|}
\hline
Embedding Layer & Model  & Embedding Dimensions   & Max Rank  \\ \thickhline
& Baseline  &  $10,131,227
\times 128$   & NA\\
& CP & [200,220,250,128] & 350  \\
1 & Tucker &  [200,220,250,128]   & 20\\
& TT &  [200,220,250,128]   & 24 \\
& TTM &  [200,220,250],[4,4,8] & 16 \\ \thickhline

& Baseline &  $2,202,608\times 128$   & NA\\
& CP & [125,130,136,128] & 306 \\
2 & Tucker &  [125,130,136,128] & 20\\
& TT &  [125,130,136,128] & 24 \\
& TTM &  [125,130,136],[4,4,8] & 16 \\ \thickhline

& Baseline &  $8,351,593\times 128$   & NA\\
& CP & [200,200,209,128] & 333 \\
3 & Tucker &  [200,200,209,128]   & 22\\
& TT &  [200,200,209,128]   & 24 \\
& TTM &  [200,220,250],[4,4,8] & 16 \\ \thickhline

& Baseline &  $5,461,306\times 128$   & NA\\
& CP & [166,175,188,128] & 326 \\
4 & Tucker &  [166,175,188,128]   & 21\\
& TT &  [166,175,188,128]   & 24 \\
& TTM &  [166,175,188],[4,4,8] & 16 \\ \thickhline

& Baseline &  $7,046,547\times 128$   & NA\\
& CP & [200,200,200] & 335 \\
5 & Tucker &  [200,200,200,128]   & 22\\
& TT &  [200,200,200,128]   & 24 \\
& TTM &  [200,200,200],[4,4,8] & 16 \\ \hline
\end{tabular}
\end{center}
}
\end{table}

%

\end{document}